\documentclass{article}

    \PassOptionsToPackage{numbers, compress}{natbib}

    \usepackage[preprint]{neurips_2020}

\usepackage[utf8]{inputenc} 
\usepackage[T1]{fontenc}    
\usepackage{hyperref}       
\usepackage{url}            
\usepackage{booktabs}       
\usepackage{amsfonts}       
\usepackage{nicefrac}       
\usepackage{microtype}      
\usepackage[pdftex]{graphicx}
\usepackage{subfig}
\usepackage{xcolor}
\usepackage{rotating}
\usepackage[ruled,vlined]{algorithm2e}

\usepackage{amsmath,amsfonts,bm}

\def\eqref#1{equation~\ref{#1}}

\def\1{\bm{1}}

\def\vtheta{{\bm{\theta}}}

\def\evtheta{{\theta}}

\DeclareMathAlphabet{\mathsfit}{\encodingdefault}{\sfdefault}{m}{sl}
\SetMathAlphabet{\mathsfit}{bold}{\encodingdefault}{\sfdefault}{bx}{n}
\newcommand{\tens}[1]{\bm{\mathsfit{#1}}}

\def\tT{{\tens{T}}}
\def\tU{{\tens{U}}}

\def\sM{{\mathbb{M}}}

\newcommand{\etens}[1]{\mathsfit{#1}}

\def\etT{{\etens{T}}}
\def\etU{{\etens{U}}}

\title{Bespoke vs. Prêt-à-Porter Lottery Tickets: Exploiting Mask Similarity for Trainable Sub-Network Finding}

\author{Michela Paganini \\
Facebook AI Research \\
\texttt{michela@fb.com} \\
\And
Jessica Zosa Forde
\\
Brown University\\
Facebook AI Research \\
\texttt{jessica\_forde@brown.edu} \\
}

\begin{document}

\maketitle

\begin{abstract}
The observation of sparse trainable sub-networks within over-parametrized networks -- also known as Lottery Tickets (LTs) -- has prompted inquiries around their trainability, scaling, uniqueness, and generalization properties. 
Across 28 combinations of image classification tasks and architectures, we discover differences in the connectivity structure of LTs found through different iterative pruning techniques, thus disproving their uniqueness and connecting emergent mask structure to the choice of pruning.
 In addition, we propose a consensus-based method for generating refined lottery tickets.
This lottery ticket denoising procedure, based on the principle that parameters that always go unpruned across different tasks more reliably identify important sub-networks, is capable of selecting a meaningful portion of the architecture in an embarrassingly parallel way, while quickly discarding extra parameters without the need for further pruning iterations. We successfully train these sub-networks to performance comparable to that of ordinary lottery tickets.
\end{abstract}

\section{Introduction}
Deep neural architectures have seen a dramatic increase in size over the years~\cite{openAIblogpost}. Over-parametrized networks exhibit high generalization performance, with recent empirical evidence showing that the generalization gap tends to close with increased number of parameters~\cite{Zhang2017No0, 2018arXiv181211118B, 2019arXiv190308560H, 2018arXiv181104918A, 2018arXiv181002054D, 2018arXiv180301206D}, contrary to prior belief, also depending on the inductive bias and propensity to memorization of each base architecture~\cite{Zhang2019-av}. 
While advantageous under this point of view, the proliferation of parameters in neural architectures may induce adverse consequences. 
For instance, the computational cost to train some state-of-the-art models has raised the barrier to entry for many researchers hoping to contribute.
Because of limited memory, time, and compute, and to enable private, secure, on-device computation, methods for model compression have seen a rise in popularity. Among these are techniques for model pruning, quantization, and distillation.

Pruning, in particular, has been seen as an overfitting avoidance method since the early decision tree literature~\cite{breiman1984classification}, with work comparing the effects of different tree pruning techniques~\cite{Mingers1989}. In analogy to this prior line of work, we provide an empirical comparison of the effects of neural network pruning methods, and analyze the structure of the sparse sub-networks that emerge.
Revived by the recent observation of the existence of sub-networks with favorable training properties within larger over-parametrized models, known as (winning) Lottery Tickets~\cite{Frankle2018-po}, and in the spirit of other work along these lines~\cite{Zhou2019-xh}, this work investigates the dependence of the properties of these sparse, trainable sub-networks on the choice of pruning technique. We set out to answer the questions: do different pruning methods identify the same lucky sub-network? If not, where are they similar? What does this tell us about fundamental pathways within neural networks, and what distinguishes a LT from other node combinations? The goal of this work is to investigate the nature of LTs and whether their structure is meaningful.

Additionally, preliminary evidence for LT transferability has been provided in both natural and non-natural image domains~\cite{2019arXiv190602773M, sabatelli2020transferability}. In the context of LTs, transfer refers to the ability to retrain a sub-network on a new task. We investigate whether similarities in the structure of LTs sourced on different datasets may, at least partially, explain their transferability across tasks.

\begin{figure}
    \centering
    \includegraphics[width=\textwidth]{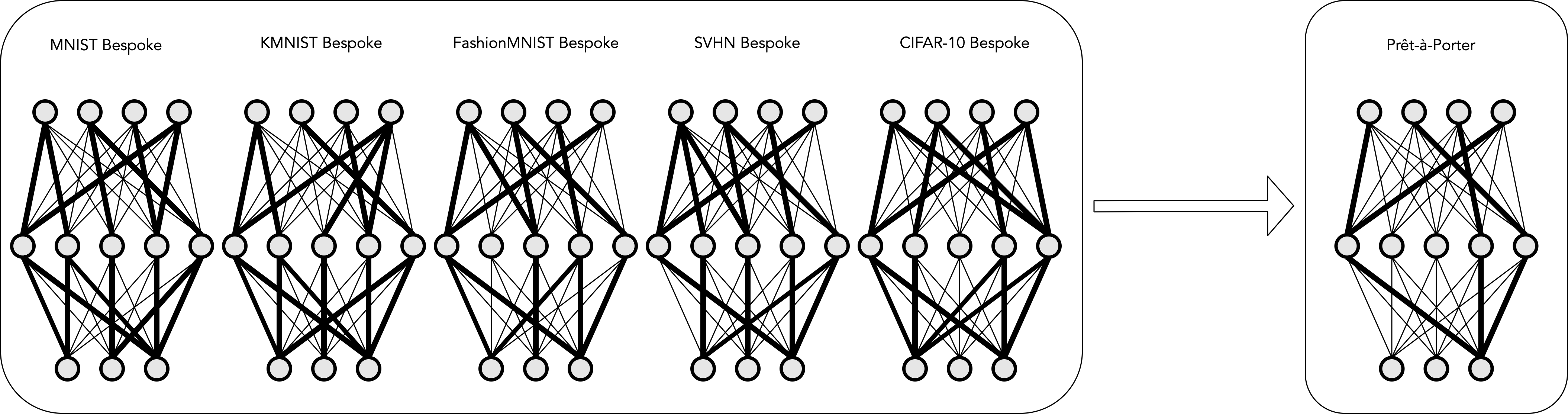}
    \caption{Consensus-based lottery ticket identification from bespoke lottery tickets sourced on different tasks. Thick lines identify unpruned connections; thin lines, pruned ones.}
    \label{fig:diagram}
\end{figure}

We exploit these similarities to denoise and further slim down the ticket, by expediting parameter removal via mask consensus pooling. Intuitively, this builds on the hypothesis that any core structure within a base architecture that consistently remains unpruned across \textit{bespoke} LTs sourced on individual tasks may more robustly identify critical pathways in the network. 
We suggest using the intersection of low-sparsity bespoke LTs to obtain a new high-sparsity LT that captures the shared, general structure needed to perform the image classification tasks considered in this work (Fig.~\ref{fig:diagram}). 
We show that the sparse structure that emerges from this procedure is itself a trainable sub-network, which we mnemonically call a \textit{prêt-à-porter} LT. This ensembling strategy also allows to parallelize the otherwise purely sequential lottery ticket finding procedure, while maintaining competitive accuracy across tasks.

\subsection{Contributions}
This work provides empirical evidence showing that:
\begin{enumerate}
    \item there can exist multiple different lucky sub-networks (lottery tickets) within an over-parametrized network;
    \item different pruning techniques are capable of finding different lottery tickets;
    \item although lottery tickets can be successfully transferred and retrained on different tasks within a family of tasks, the masks sourced on different tasks are largely dissimilar;
    \item the intersection of masks sourced on different tasks identifies a new, sparser, trainable lottery ticket that allows to parallelize the sparsification procedure through a more grounded node importance definition.
\end{enumerate}

\section{Related work}
\label{sec:related_work}
State-of-the-art results in machine learning seem to strongly correlate with increased number of parameters~\cite{brown2020language}, leading to the common use of vastly over-parametrized models, compared to the complexity of the tasks at hand.  
In practice, however, most common neural network architectures can be dramatically pruned down in size, though there is no natural order in which neural network parameters should be removed~\cite{hanson1989comparing, LeCun1990-qd, NIPS1992_647, 2015arXiv151000149H, 2016arXiv160804493G, 2016arXiv161105128Y, 2017arXiv170706342L, 2017arXiv170709102T, Zhang_2018_ECCV, AYINDE2019148, 2019arXiv190704018M}. 

Identifying important units in a neural network is an open challenge, often driven by desires for interpretability and model compression, among others. Various proxies for importance have been proposed as a direct consequence of the lack of precise mathematical definition of this term. Different objectives lead to different mathematical formulations of importance: concept creation, output attribution, conductance, saliency, activation, explainability, connectivity density, redundancy, noise robustness, stability, and more~\cite{bmmutual, Amjad2018-do, Zhou2018-xy, 2018arXiv180512233D, shrikumar2017learning}.

In general, though, small, sparse models are empirically hard to train from scratch. 
However, the recently formulated lottery ticket hypothesis conjectures that there exists at least one sparse sub-network, a \textit{winning ticket}, within an over-parametrized network, that can achieve commensurate accuracy in commensurate training time with fewer parameters, if retrained from the same initialization~\cite{Frankle2018-po}. To find this lucky sub-network, the lowest magnitude connections are iteratively pruned for a fixed number of iterations~\cite{LeCun1990-qd, Han2015-ze}. Interestingly, while pruned weights are set to zero, unpruned weights are \textit{rewound} to their initialization values, prior to the next iteration of training and pruning.

Researchers have reported difficulties in scaling this method to larger models. Some attempts were unable to successfully train small sparse sub-networks found through this procedure, when starting from VGG-16~\cite{Simonyan2014-lt} and ResNet-50~\cite{He2015-jb} trained on CIFAR-10~\cite{Krizhevsky2009-fx}, and instead observed that random initialization outperformed weight rewinding~\cite{Liu2018-lc}. 
Similarly, other experimental work was unable to successfully produce a sub-network with these properties within ResNet-50 trained on ImageNet~\cite{Russakovsky2015-ib} and Transformer~\cite{Vaswani2017-iu} trained on WMT 2014 English-to-German~\cite{Gale2019-xx}.\footnote{Translation Task - ACL 2014 Ninth Workshop on Statistical Machine Translation (URL: \url{https://www.statmt.org/wmt14/translation-task.html})}
Modifications to the original procedure have been proposed in order to favor the emergence of winning tickets in larger models and tasks. These include challenging the need for unstructured, magnitude-based pruning~\cite{Zhou2019-xh}, using learning rate warmup~\cite{Frankle2018-po}, selecting unimportant weights globally instead of layer by layer~\cite{Frankle2018-po}, and \textit{late resetting} unpruned parameters to values achieved early in training, as opposed to rewinding the weights all the way to their initial values~\cite{Frankle2019-vu, 2019arXiv190602773M}. These strategies were further utilized to show that lottery tickets are not simply an emergent phenomenon in the context of supervised learning of natural images, but they can be found and trained on NLP and RL tasks as well~\cite{yu2019playing}. This debate has left questions unanswered around the nature of LTs, the significance of their structure and initialization, their implications for information propagation and concept creation, and their uniqueness within over-parametrized architectures. Furthermore, the lack of principled understanding of what distinguishes a LT from other sub-networks within the same model has made it hard to move beyond the computationally expensive, iterative LT-finding technique based on magnitude-based unstructured pruning.

On the other hand, lottery tickets have been shown to possess generalization properties that allow for their reuse across similar tasks, thus reducing the computational cost of finding dataset-dependent sparse sub-networks~\cite{2019arXiv190602773M, sabatelli2020transferability}. However, the mechanism that powers their transferability properties is not yet clearly understood, and is therefore subject to further study in this work.

Alternative approaches to neural architecture optimization have taken a constructionist approach, starting from a minimal set of units and connections, and growing the network by adding new components~\cite{NEAT, 2019arXiv190401569X}.
Finally, other methods combine both network growing and pruning, in analogy to the different phases of connection formation and suppression in the human brain~\cite{Floreano2008, dai2019nest}.

\section{Methodology}
\label{sec:methodology}
We explore the performance of sub-networks generated by different iterative pruning techniques starting from base LeNet~\cite{LeCun1990-em}, AlexNet~\cite{alexnet}, VGG11~\cite{Simonyan2014-lt}, and ResNet18~\cite{He2015-jb} architectures, on the following set of image classification tasks: MNIST~\cite{mnist:1994}, Fashion-MNIST~\cite{fashionmnist}, Kuzushiji-MNIST (or KMNIST)~\cite{clanuwat2018deep, kmnist}, EMNIST~\cite{cohen2017emnist}, CIFAR-10, CIFAR-100~\cite{Krizhevsky2009-fx}, SVHN~\cite{netzer2011reading}. 

All networks are trained for 30 epochs using SGD with constant learning rate 0.01, batch size of 32, without explicit regularization. The pruning fraction per iteration is held constant at 20\% of remaining connections/units per layer.

In all figures, unless otherwise specified, the error bars and shaded envelopes correspond to one standard deviation (half up, half down) from the mean, over 6 experiments with seeds 0-5.

\subsection{Pruning methods}
\label{sec:pruning}
This works explores a variety of pruning techniques that may differ along the following axes:

\textbf{Neuronal importance definition}: In magnitude-based pruning, units/connections are removed based on the magnitude of synaptic weights. Usually, low magnitude parameters are removed. As a (rarer) alternative, one can consider removing high magnitude weights instead~\cite{Zhou2019-xh}.
Non-magnitude-based pruning techniques can be based, among others, on activations, gradients, or custom rules
for neuronal importance.

\textbf{Local vs. global}:
Local pruning consists of removing a fixed percentage of units/connections from each layer by comparing each unit/connection exclusively to the other units/connections in the layer. On the contrary, global pruning pools all parameters together across layers and selects a global fraction of them to prune. The latter is particularly beneficial in the presence of layers with unequal parameter distribution, by redistributing the pruning load more equitably. A middle-ground approach is to pool together only parameters belonging to layers of the same kind, to avoid mixing, say, convolutional and fully-connected layers.

\textbf{Unstructured vs. structured}:
Unstructured pruning removes individual connections, while structured pruning removes entire units or channels. Note that structured pruning along the input axis is conceptually similar to input feature importance selection.
Similarly, structured pruning along the output axis is analogous to output suppression.

In this work, we compare: magnitude-based \{global, $L_1$, random\} unstructured, and \{$L_1$, $L_2$, $L_{-\infty}$, random\} structured pruning. 
Pruning is only applied to weight matrices and not biases (nor other parametrized layers such as batch norm~\cite{ioffe2015batch}).
Pruning techniques and models are implemented using PyTorch~\cite{paszke2017automatic, paganini2020streamlining}.

\section{Analysis}

\begin{figure}
\centering
\subfloat[VGG11 on CIFAR-100]{\includegraphics[width=0.23\textwidth]{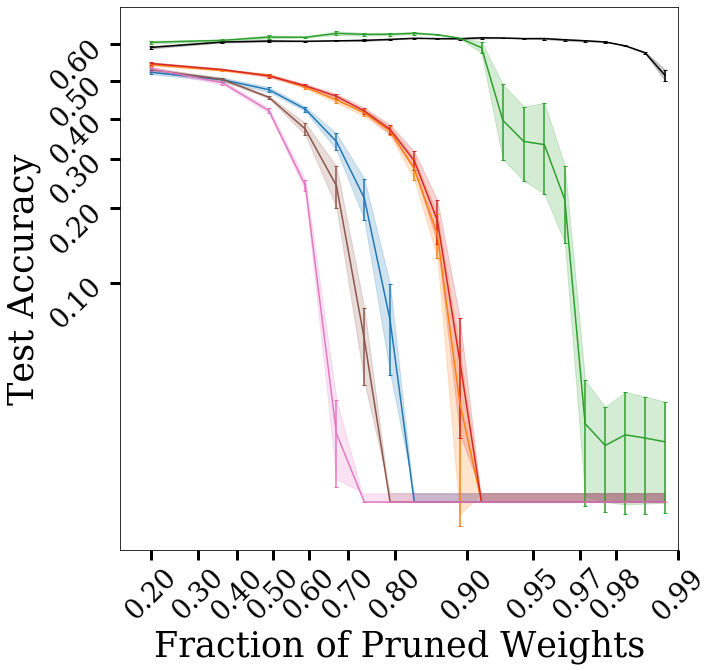}\label{fig:cifar100_vgg11}}
\hspace{0.2cm}
  \subfloat[AlexNet on CIFAR-100]{\includegraphics[width=0.23\textwidth]{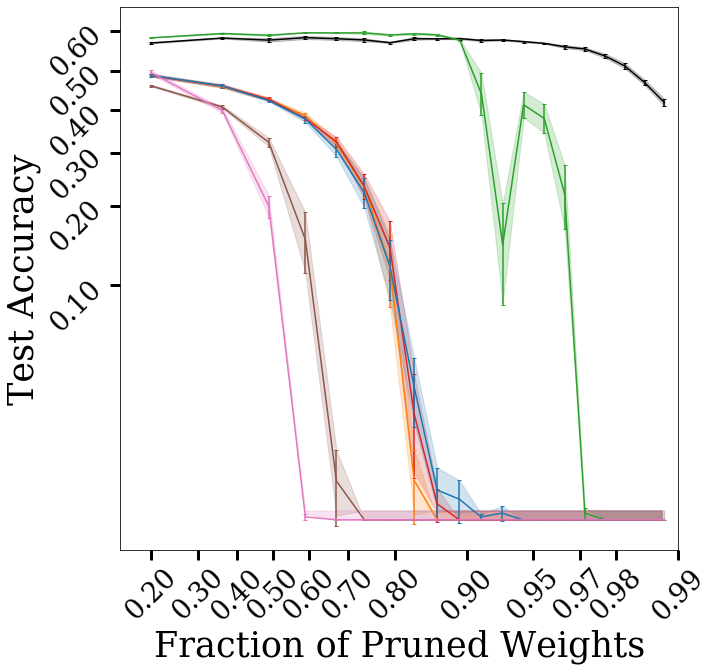}\label{fig:cifar100_alexnet}}
  \hspace{0.2cm}
 \subfloat[ResNet18 on CIFAR-10]{\includegraphics[width=0.23\textwidth]{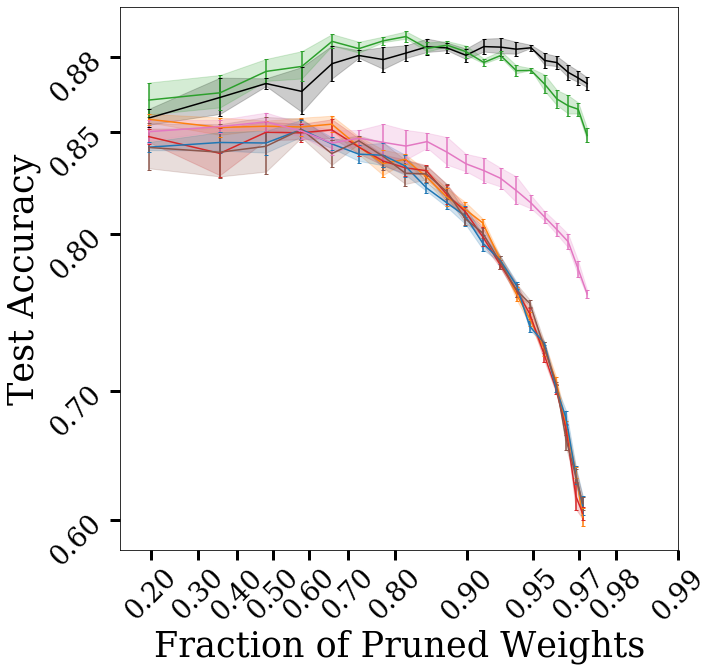}\label{fig:cifar-10_resnet}}
 \hspace{0.2cm}
  \subfloat[LeNet on FashionMNIST]{\includegraphics[width=0.23\textwidth]{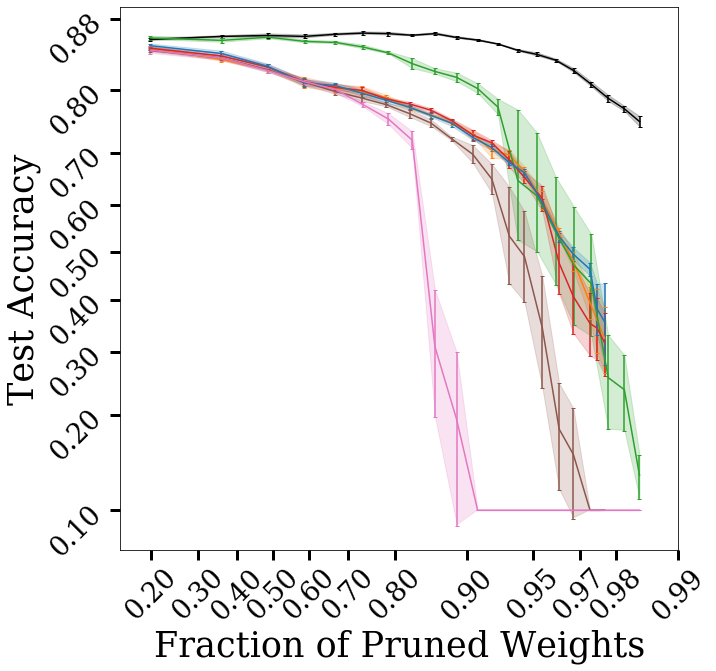}\label{fig:fmnist_lenet}}
   \caption{Test accuracy for SGD-trained, sparsified models, pruned using seven different pruning techniques (\textcolor{red}{$L_2$-structured}, global unstructured, \textcolor{green}{$L_1$-unstructured}, \textcolor{orange}{$L_1$-structured}, \textcolor{blue}{$L_\infty$-structured}, \textcolor{brown}{random structured}, \textcolor{pink}{random unstructured}), and rewound to initial weight values after each pruning iteration.}\label{fig:accspars}
\end{figure}

Each point on the accuracy-sparsity trade-off curves in Fig.~\ref{fig:accspars} represents the test accuracy after exactly 30 epochs of training of a pruned model at the given level of sparsity. In these experiments, models are trained and evaluated on the same task, throughout the iterative sparsification procedure.
Both global and local unstructured pruning are able to identify high-performance sub-networks that are trainable up to high levels of sparsity. The additional flexibility of global pruning to compare weight magnitude across all layers, instead of in a layer-by-layer manner, allows it to prune larger, more over-parametrized layers more aggressively, while keeping other layers almost intact.
Performance plots for other dataset-model combinations can be found in Appendix~\ref{app:accspars}. 

\begin{figure}
\centering
  \subfloat{\includegraphics[width=0.25\textwidth]{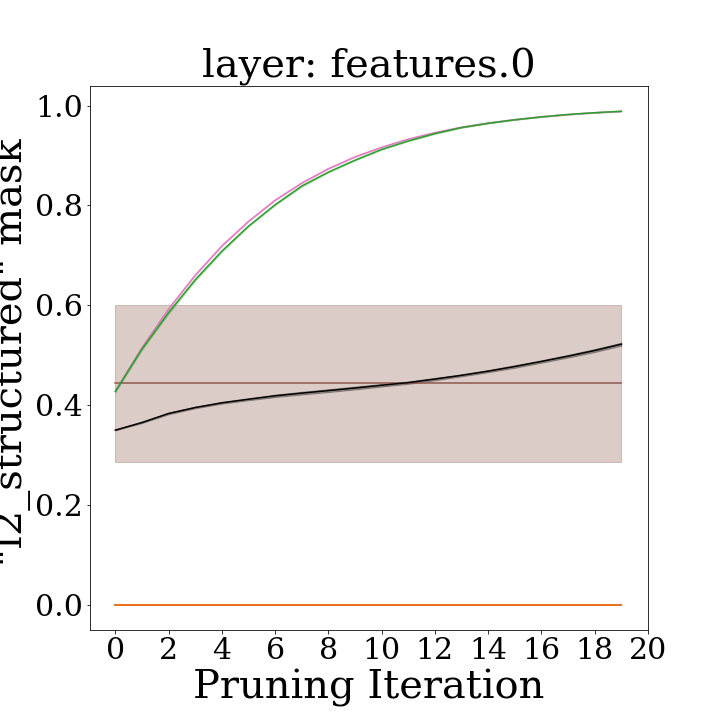}}
  \subfloat{\includegraphics[width=0.25\textwidth]{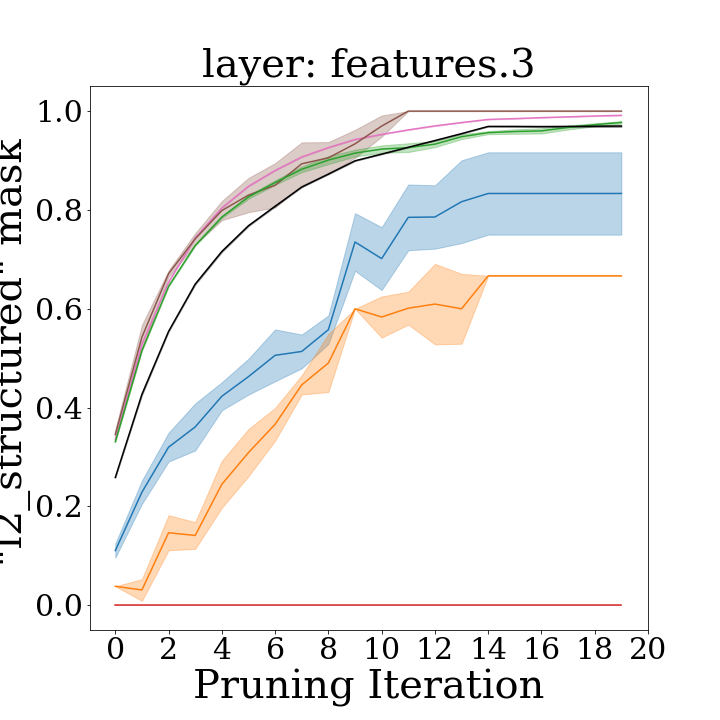}}
  \subfloat{\includegraphics[width=0.25\textwidth]{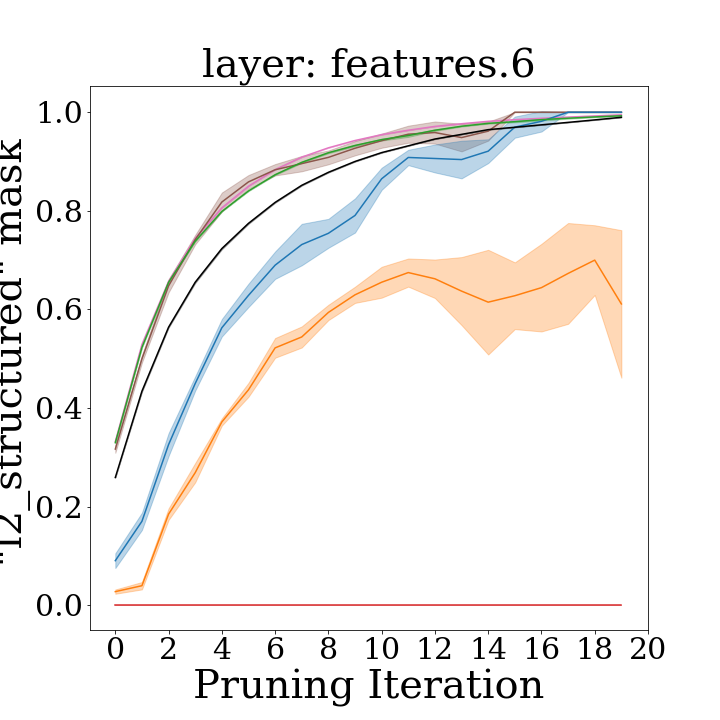}}
  \subfloat{\includegraphics[width=0.25\textwidth]{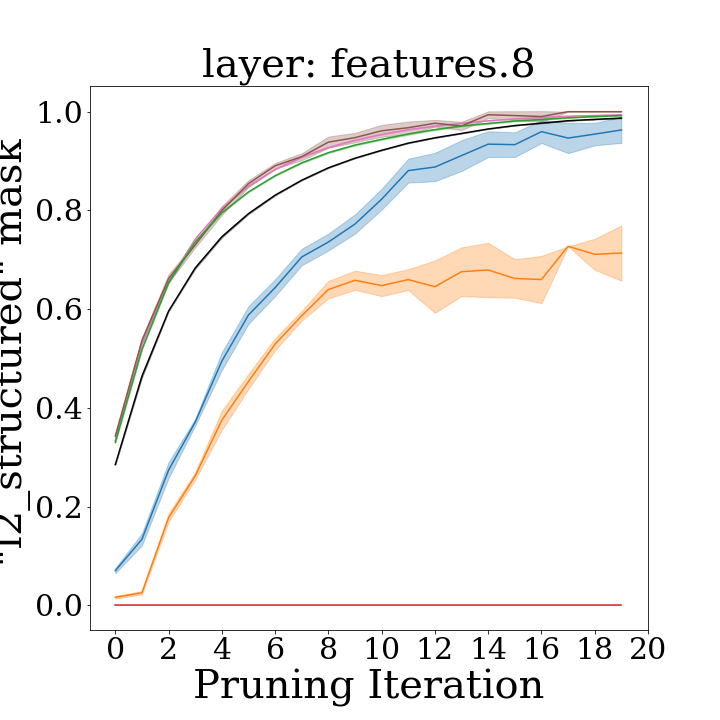}}\\
  \subfloat{\includegraphics[width=0.25\textwidth]{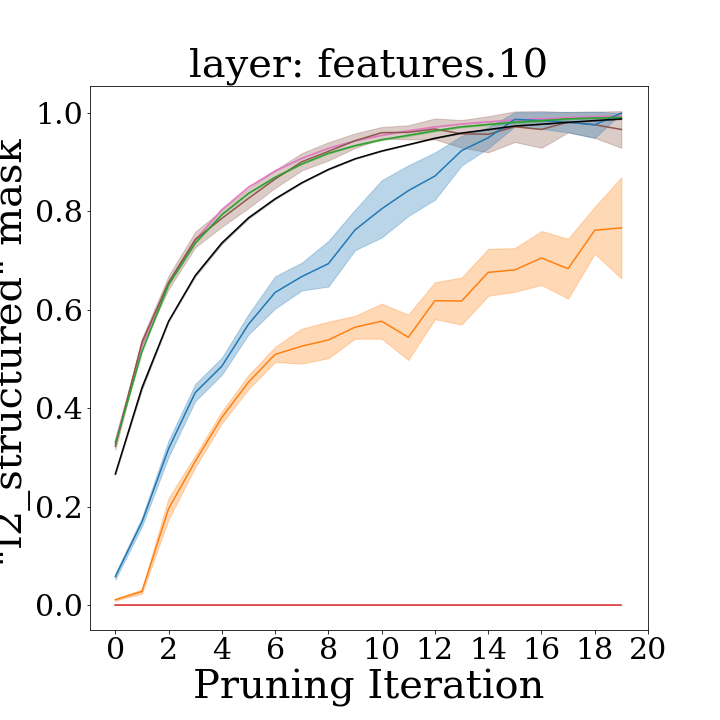}}
  \subfloat{\includegraphics[width=0.25\textwidth]{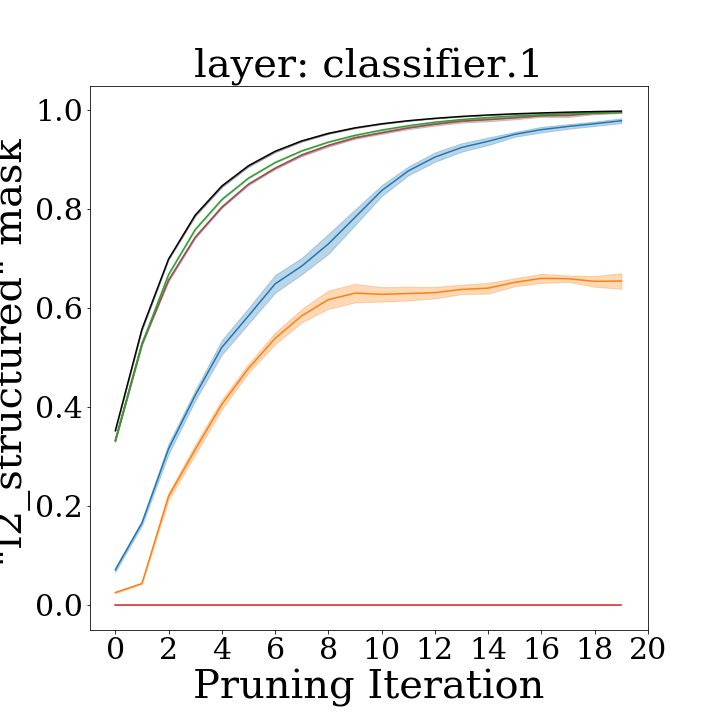}}
  \subfloat{\includegraphics[width=0.25\textwidth]{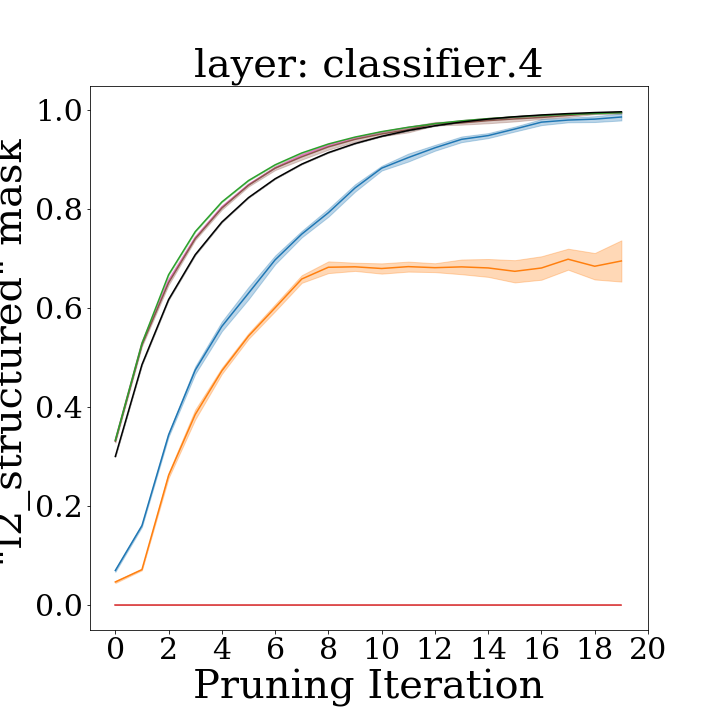}}
  \subfloat{\includegraphics[width=0.25\textwidth]{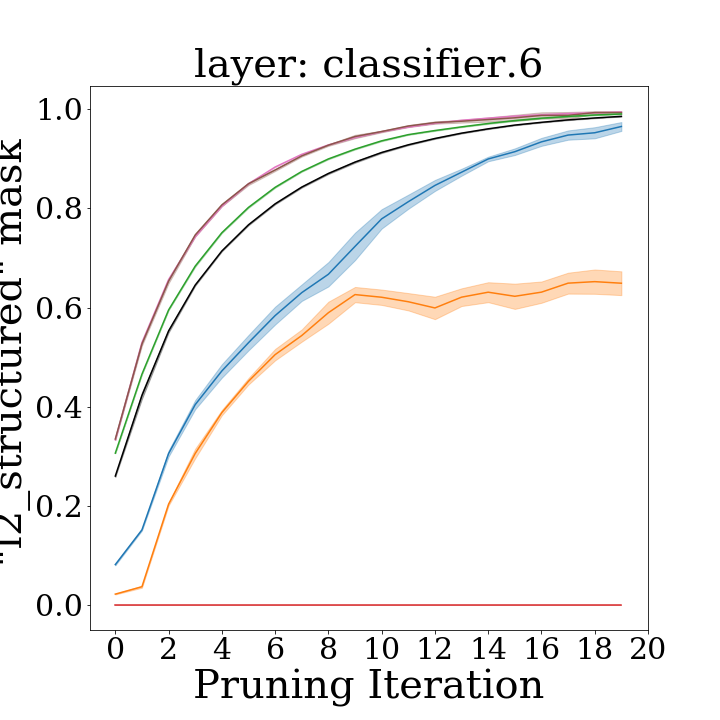}}
   \caption{Layer-wise Jaccard distance between masks found by pruning AlexNet on CIFAR-100 using \textcolor{red}{$L_2$-structured} pruning, and masks found by other pruning methods (global unstructured, \textcolor{green}{$L_1$-unstructured}, \textcolor{orange}{$L_2$-structured} \textcolor{blue}{$L_\infty$-structured}, \textcolor{brown}{random structured}, \textcolor{pink}{random unstructured}), as a function of the pruning iteration, conditional on identical seed, across six seeds. Each plot represents a layer in AlexNet. $L_1$-structured pruning yields the most similar masks to $L_2$-structured pruning, as expected.}\label{fig:jaccard}
\end{figure}

Different pruning techniques are therefore capable of identifying different sparse trainable sub-networks, although, based on pure accuracy-sparsity trade-off arguments, some may be able to extract more meaningful structure than others. 

\subsection{The similarity of masks}

In what follows, let a mask $\sM$ with $|\sM| = D$ over a network of $D$ parameters $\vtheta$ be a set of elements with binary values $m_i = {\bm{1}}_\mathrm{unpruned}(\evtheta_i) \in \{0, 1\}$ that represent membership of the network's parameters to the set of unpruned weights. 
We measure the overlap between two binary masks $(\sM_1, \sM_2)$ representing the sparsification of two networks that share identical base architecture and initialization, by computing their Intersection over Union (IoU), also quantifiable in terms of the Jaccard distance~\cite{jaccard-index}: $d_J(\sM_1, \sM_2) = 1 - \frac{|\sM_1 \cap \sM_2|}{|\sM_1 \cup \sM_2|}$.

\subsubsection{Mask similarity across pruning methods}
\begin{figure}
\centering
  \subfloat[Pruning iteration 1]{\includegraphics[width=0.32\textwidth]{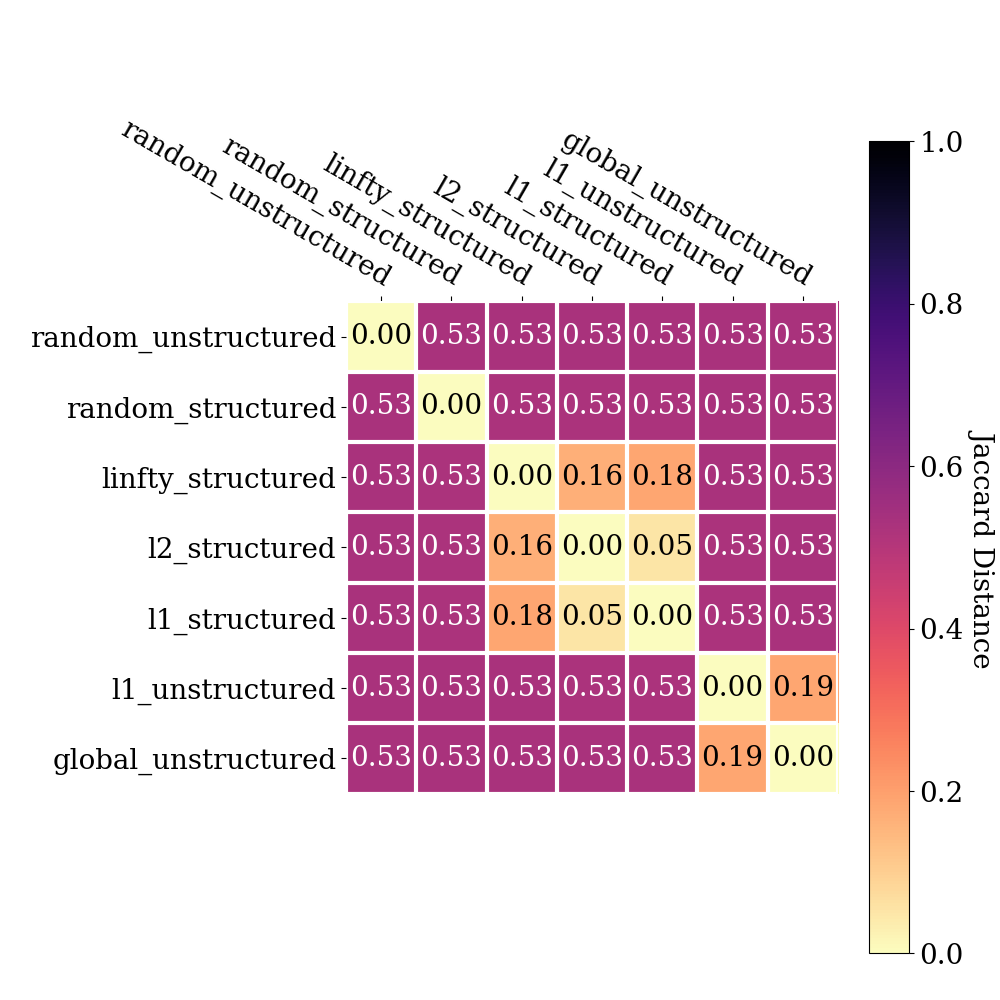}}
  \hspace{0.3em}
  \subfloat[Pruning iteration 6]{\includegraphics[width=0.32\textwidth]{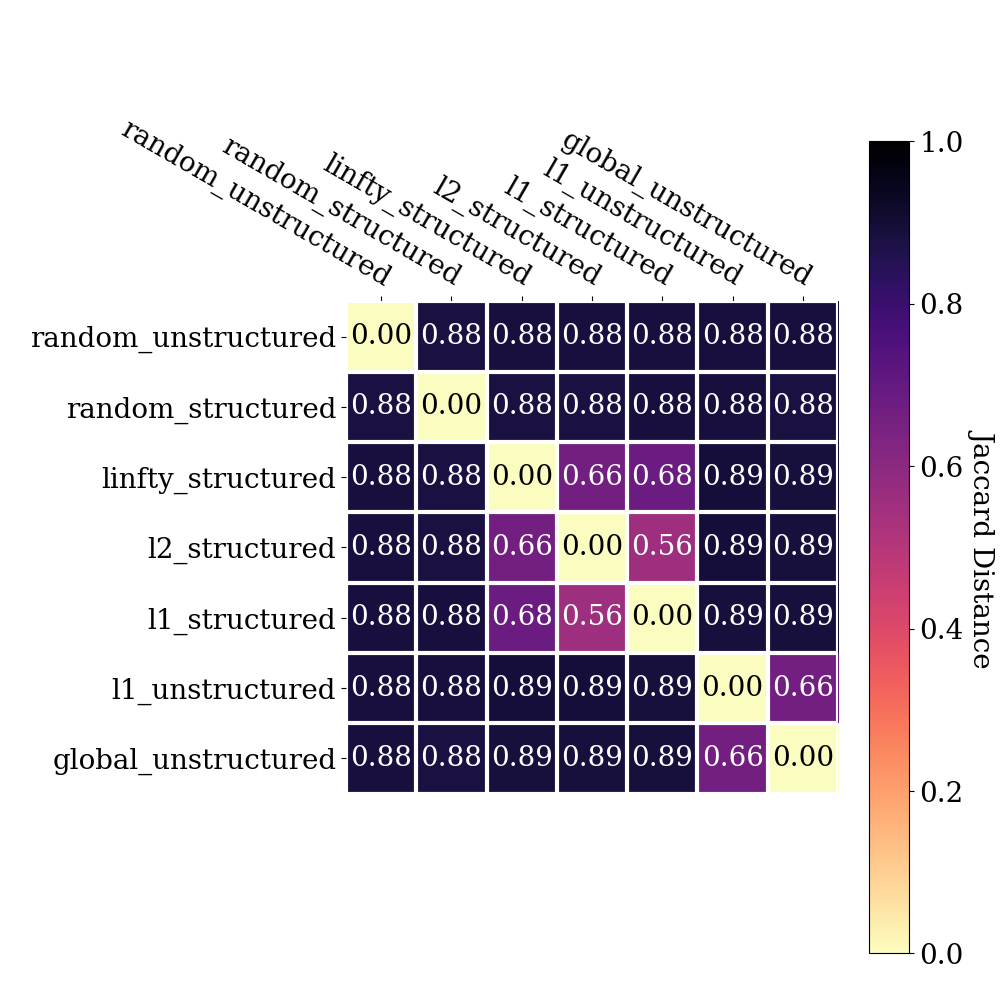}}
    \hspace{0.3em}
  \subfloat[Pruning iteration 19]{\includegraphics[width=0.32\textwidth]{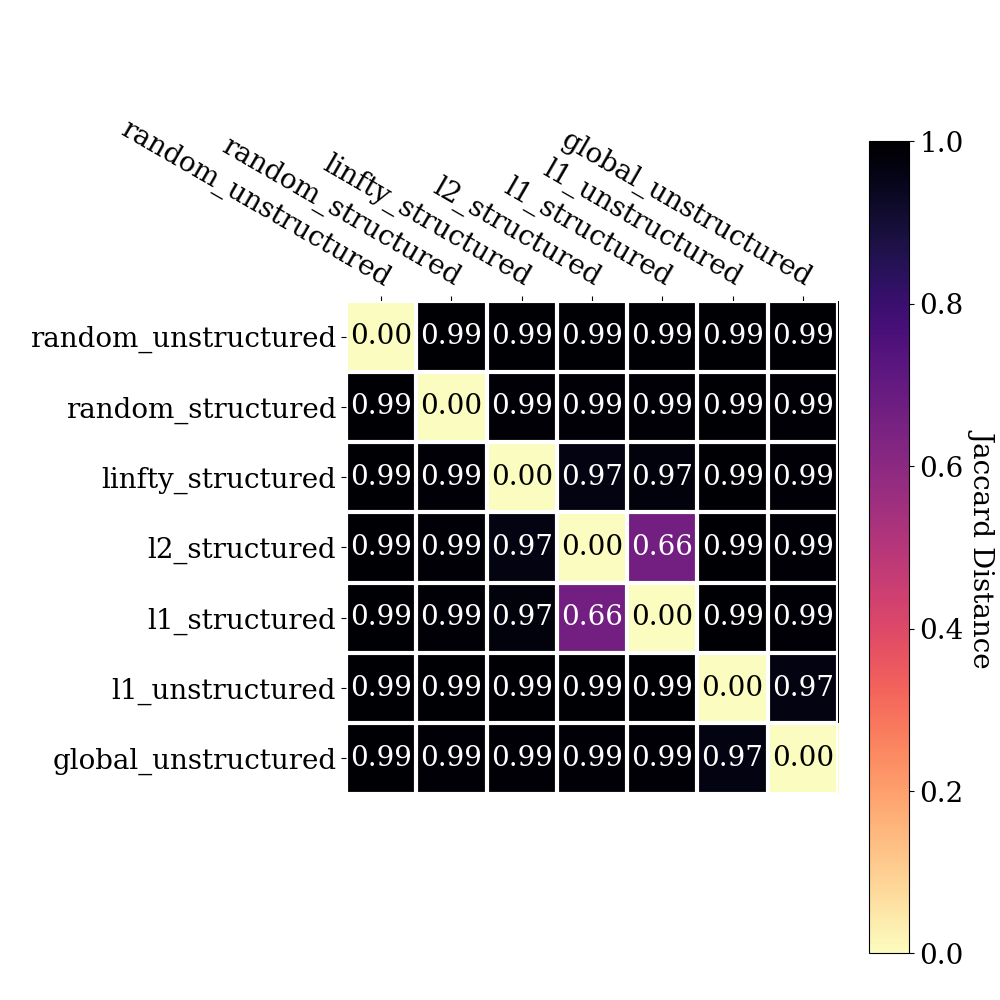}}
  \caption{Pairwise total Jaccard distance between the masks obtained by different pruning techniques on AlexNet architectures trained on CIFAR-100. As the networks grow progressively sparser, the distance between masks grows as their intersection shrinks.
  }\label{fig:jaccard_heats_prun}
\end{figure}

Although different pruning techniques may yield sub-networks with comparable total accuracy at a given sparsity level, a deeper investigation into their connectivity structure shows, as evident from the growth of the per-layer Jaccard distance as a function of pruning iteration in Fig.~\ref{fig:jaccard} and the evolution of the pairwise total Jaccard distances in Fig.~\ref{fig:jaccard_heats_prun}, that there exist multiple lucky sub-networks with similar performance, yet little to no overlap. 

Pruning the same network using different pruning techniques gives rise to sparse sub-networks that differ not only in structure but also in the learned function that they compute. Given sufficient compute and memory budget, one can consider ensembling the LTs and combining the predictions made by each sub-network to boost performance.

\subsubsection{Mask similarity across tasks}
\label{sec:jaccard_tasks}
Fixing the pruning method and initializing identical replicas of the same model, we find lottery tickets on a set of image classification tasks. We call lottery tickets sourced on a specific task \textit{bespoke} lottery tickets.  We confirm previous findings of ticket transferability ~\cite{Morcos2018-vj, sabatelli2020transferability}, and we aim to analyze whether similarity of masks can trivially explain this property.

The performance of bespoke LTs (of ${\sim}85\%$ sparsity for LeNet, and ${\sim}75\%$ sparsity on AlexNet, VGG11, and ResNet18) evaluated on a set of target tasks (Fig.~\ref{fig:across_tasks}) confirms that these LTs are indeed sufficiently general to allow for retraining on various datasets from the same domain. Each bespoke LT performs best on the task it was sourced on, but tickets sourced on other tasks generally remain competitive.

\begin{figure}
\centering
  \subfloat[Pruning iteration 2]{\includegraphics[width=0.32\textwidth]{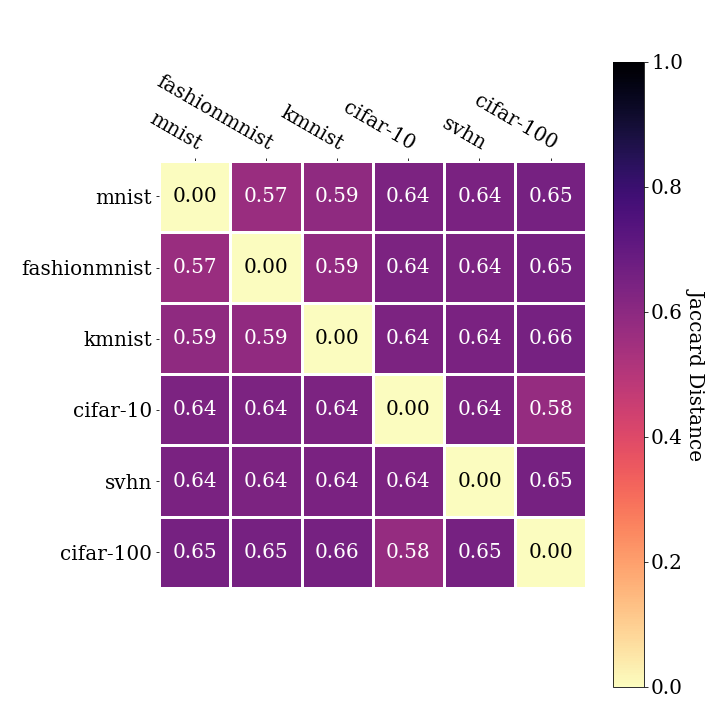}}
  \hspace{0.3em}
  \subfloat[Pruning iteration 8]{\includegraphics[width=0.32\textwidth]{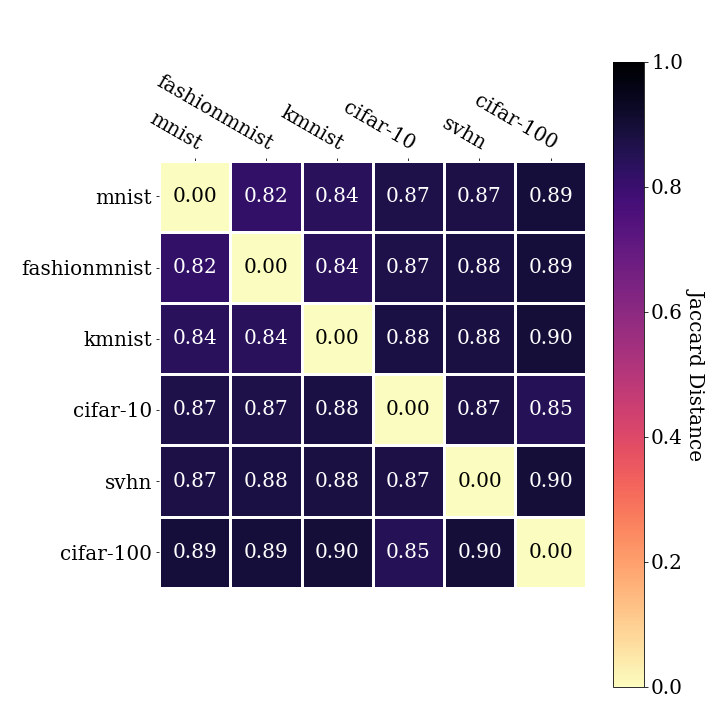}}
    \hspace{0.3em}
  \subfloat[Pruning iteration 19]{\includegraphics[width=0.32\textwidth]{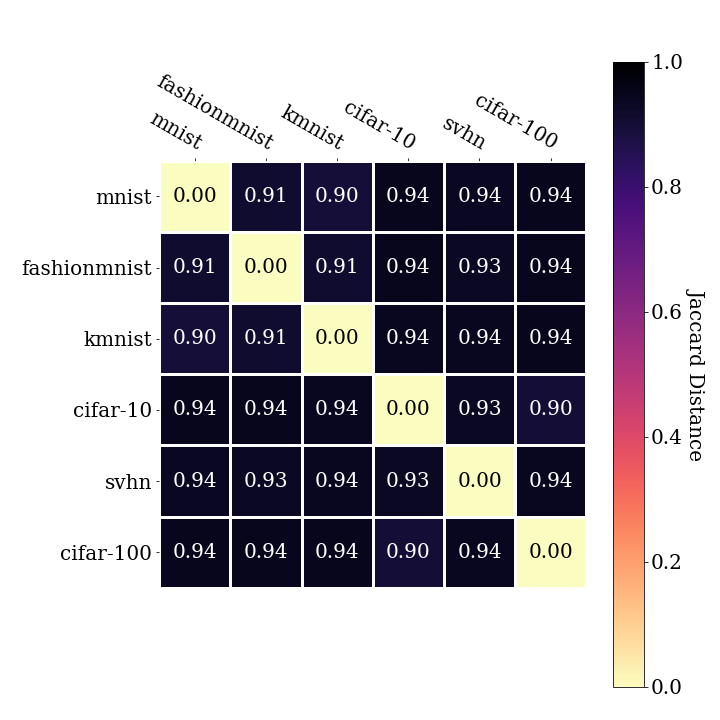}\label{fig:jaccard_lenet_tasks_19}}
  \caption{Pairwise total Jaccard distance between the bespoke masks obtained through global unstructured pruning on LeNet, over a set of tasks. As the networks grow progressively sparser, the distance between masks grows as their intersection shrinks.}\label{fig:jaccard_lenet_tasks}
\end{figure}

The average Jaccard distance between bespoke LT masks reveals significant dissimilarities between sub-networks of similar sparsity sourced on different datasets, even when data distributions are close, as is the case for CIFAR-10 and CIFAR-100, or MNIST and KMNIST. However, even at high levels of sparsity (${\sim}99\%$ in Fig.~\ref{fig:jaccard_lenet_tasks_19}), emerging masks are not completely disjoint, but share 6-10\% of their unpruned parameters with other bespoke masks.

We conclude that LT transfer capabilities cannot be trivially explained by close-to-identical bespoke masks, but these masks do share a core backbone of unpruned parameters, even at high levels of sparsity.

\subsection{Prêt-à-porter lottery tickets: transferable, shared, core sub-networks}

The LT finding algorithm relies on a noisy neuron importance ranking based on absolute weight magnitude, which has non-negligible probability of mischaracterizing and, therefore, eliminating important parameters and pathways in a network. For this reason, traditional LT-style pruning proceeds cautiously through several computationally expensive iterations of pruning, rewinding, and retraining, by only removing a low fraction (20\% being an effective compromise) of remaining weights at each iteration. In fact, while magnitude-based weight pruning has both theoretically and experimentally grounded reasons for being one of the preferred proxies for neuron importance~\cite{Han2015-ze, Gale2019-xx}, factors such as the stochasticity in the training process, the underconstrainedness of weights in over-parametrized networks (especially in ReLU networks), the non-immediate rate of convergence of parameters to their final value, as well as any other source of noise that enters the final magnitude-based ranking of weights, make this proxy unstable and the lottery ticket discovery procedure noise prone.

Exploiting the insight from Sec.~\ref{sec:jaccard_tasks} regarding the existence of shared portions of LT structure across tasks, we propose a method for eliminating unwanted noise in the lottery ticket structure by relying on consensus-based refinement of low-to-medium sparsity LTs into higher sparsity LTs that we call \textit{prêt-à-porter} lottery tickets (Fig.~\ref{fig:diagram}). We believe this to be a computationally efficient, logically compelling, and principled way of identifying key parameters within over-parametrized networks, that can rival multiple rounds of iterative magnitude-based pruning for LT discovery.

Given a number of tasks $N$, a consensus mask $\sM$ is computed so that each entry represents the fraction of tasks over which, according to the traditional LT-finding algorithm, the corresponding weight goes unpruned. 
If the traditional LT-finding algorithm consistently picked out the same sub-network, independent of the task, then the consensus mask $\sM$ would be a binary mask identical to the bespoke LT masks. The opposite limit would be the case of no overlap across masks. Given the partial similarities observed among bespoke masks sourced on different datasets, one can normally expect a non-binary $\sM$.
In this work, we demand that weights be unpruned across all $N$ bespoke tickets to become members of the prêt-à-porter ticket: $\sM = \sM_1 \cap \sM_2 \cap \sM_3 \cap \cdots \cap \sM_N$. A different threshold can be imposed to relax this notion and obtain prêt-à-porter tickets with lower sparsity.
This intersection extracts the core, shared network structure that is deemed important for efficiently solving multiple tasks. 

The computation of the low-sparsity LTs $\sM_i$ used to obtain the mask intersection $\sM$ can be made embarrassingly parallel, thus addressing one of the major practical obstacles for LT adoption.

\subsubsection{Experiments}
We consider a set of $N=5$ tasks from the same domain (in this case, image classification). We select tasks that share the same number of output classes, here $k=10$, and we resize the input images to the same shape, in order to ensure identical base architecture. The datasets used in this portion are: MNIST, FashionMNIST, KMNIST, CIFAR-10, SVHN.
A copy of the model (with identical seed, thus identical initialization) is trained on each separate task, then iteratively pruned with weight rewinding over 2 pruning iterations, using global unstructured pruning with 20\% sparsification rate per iteration. The training hyper-parameters are identical to the ones in Sec.~\ref{sec:methodology}.

The binary masks of the resulting bespoke LTs are queried for consensus, in order to identify core weights that consistently go unpruned across the full set of $N=5$ tasks. We expect the masks' intersection to represent fundamental pathways within the base architecture that are important for solving this family of tasks.

The resulting prêt-à-porter ticket has a mask equal to the intersection of masks found across the $N=5$ tasks, and initialization equal to the value at initialization of those remaining weights. The prêt-à-porter and all bespoke tickets are retrained from scratch on each target task. 
We empirically demonstrate that the prêt-à-porter LT is, indeed, trainable to comparable performance to similar-sparsity bespoke LTs found through the traditional iterative pruning strategy (Fig.~\ref{fig:across_tasks}). For equivalent LT performance evaluation at initialization, prior to training, see Appendix~\ref{app:atinit}. The failures in training at times observed on SVHN can be remedied with a more principled training recipe; however, to keep results consistent, we refrain from dataset-specific optimization.

\begin{figure}
    \centering
    \includegraphics[width=\textwidth]{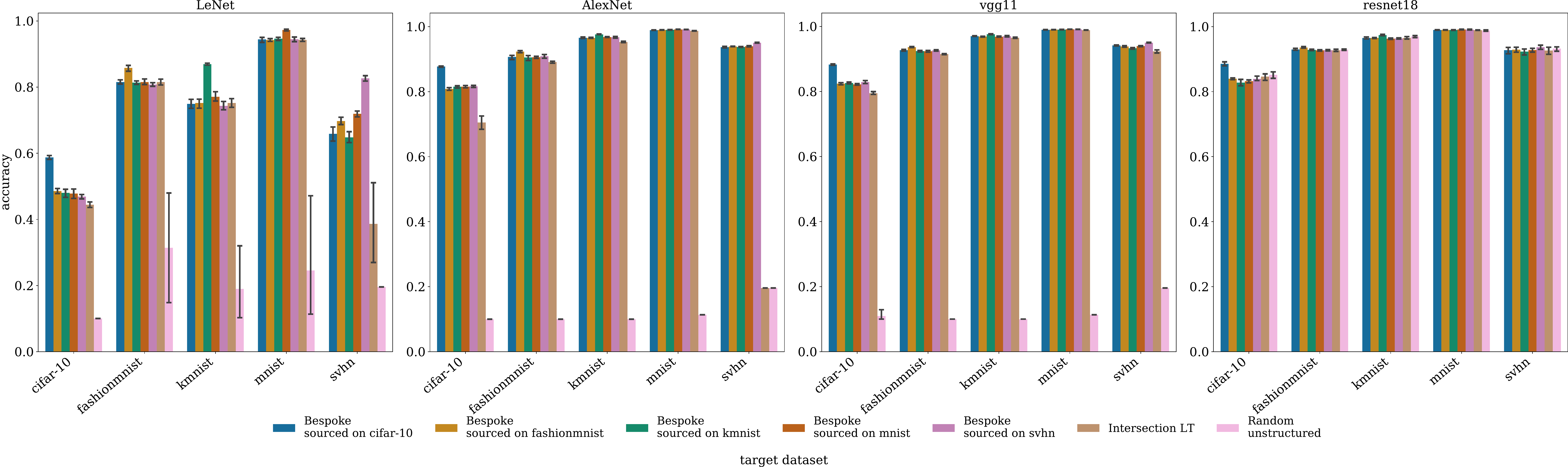}
    \caption{Bespoke and prêt-à-porter lottery ticket accuracy for networks sourced on the tasks listed in the legend using global unstructured pruning, starting from LeNet (panel 1), AlexNet (panel 2), VGG11 (panel 3), and ResNet18 architectures, and evaluated on the target tasks listed along the x-axis. As a benchmark, we also show the performance of a random sub-network. All sub-networks have comparable sparsity.}
    \label{fig:across_tasks}
\end{figure}

Furthermore, prêt-à-porter LTs have the added advantage of better learnability, as the parameters that compose them tend to be initialized from a distribution of smaller variance, closer to that of the original initialization distribution, than bespoke tickets (Fig.~\ref{fig:pap_init_alex} for AlexNet; see Appendix~\ref{appendix:gen-pap} for ResNet18, VGG11, and LeNet in Figs.~\ref{fig:pap_init},~\ref{fig:pap_init_vgg}, and~\ref{fig:pap_init_lenet} respectively). The scale of the standard deviation at initialization is known to be important for network training stability~\cite{he2015delving}.

\begin{figure}
    \centering
    \includegraphics[width=\textwidth]{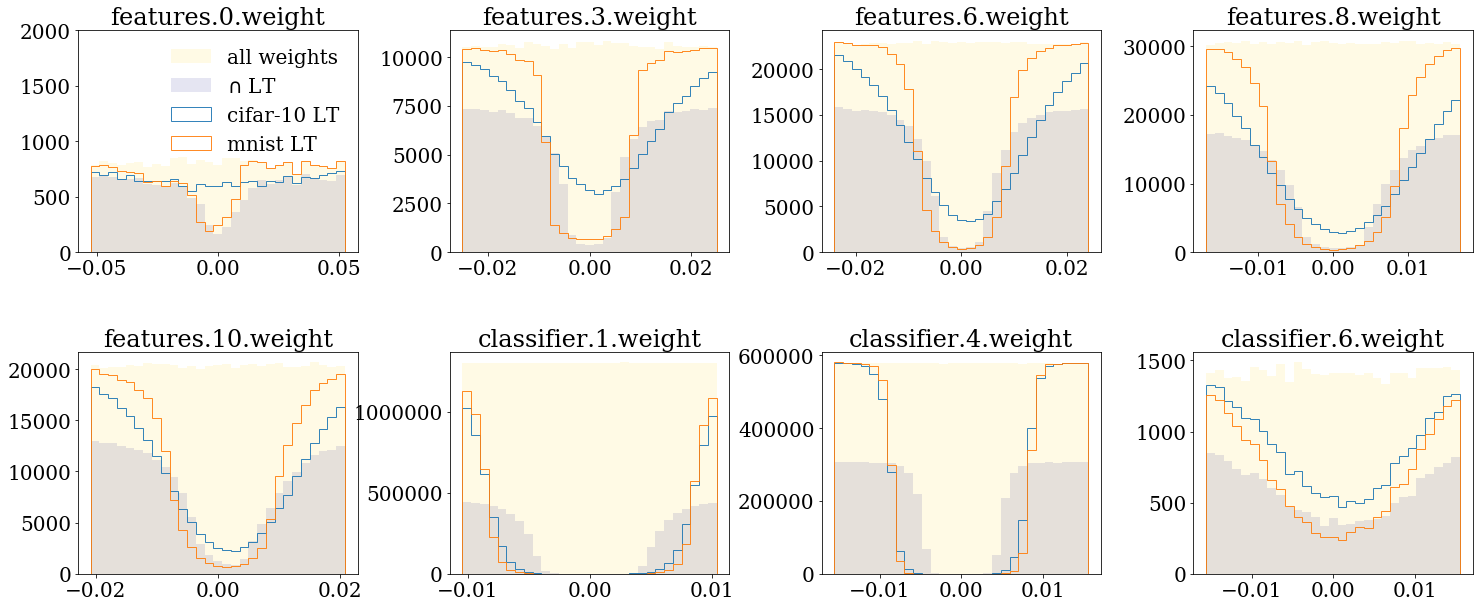}
     \caption{Initialization values of weights in a AlexNet prêt-à-porter LT (denoted $\cap \mathrm{LT}$), compared to those that make up a bespoke LT of similar sparsity sourced on CIFAR-10, and those that make up a bespoke LT of similar sparsity sourced on MNIST. For comparison, the full distribution of initial weights is shown as a shaded gold histogram.}
    \label{fig:pap_init_alex}
\end{figure}

\section{Conclusion}
We show evidence against the uniqueness of winning tickets in a variety of networks and tasks, by identifying different lucky sub-networks of competitive performance within the same parent network, while controlling for degeneracy by fixing experimental seeds. We also show that LTs initiated from a common initialization, but trained on different datasets, tend to have a small but consistent overlap of masks. Using this, we introduce \textit{prêt-à-porter} LTs, effectively combining the masks from identical models with LTs sourced on different tasks. We show that prêt-à-porter LTs are not only competitive, but provide sourcing speedups via parallelism.

We hope these experimental results in reproducible, controlled settings will guide theoretical work in this sub-field, and lead to novel approaches to LT sourcing, and more broadly, a deeper understanding of LTs and pruning as a whole.

\clearpage
\section*{Broader Impact}

This work builds upon prior work in LTs and pruning more broadly, which, beyond as a probe into fundamental questions of network capacity, aid in making neural networks more accessible in low-power or resource-constrained environments, which has massive implications for the accessibility of models in the developing world, medical applications, and ubiquity on mobile devices, to name a few. We must be mindful, however, that invasive, network modifying actions may have unintended demographic consequences with respect to fairness, bias, and more broadly reliability and interpretability -- critical corner stones when considering the environments that LTs and pruned models may be used in.

\bibliography{references}
\bibliographystyle{plainurl}

\clearpage
\appendix
\section{Additional accuracy-sparsity curves}
\label{app:accspars}
\begin{figure}
\centering
  \subfloat[AlexNet on CIFAR-10]{\includegraphics[width=0.3\textwidth]{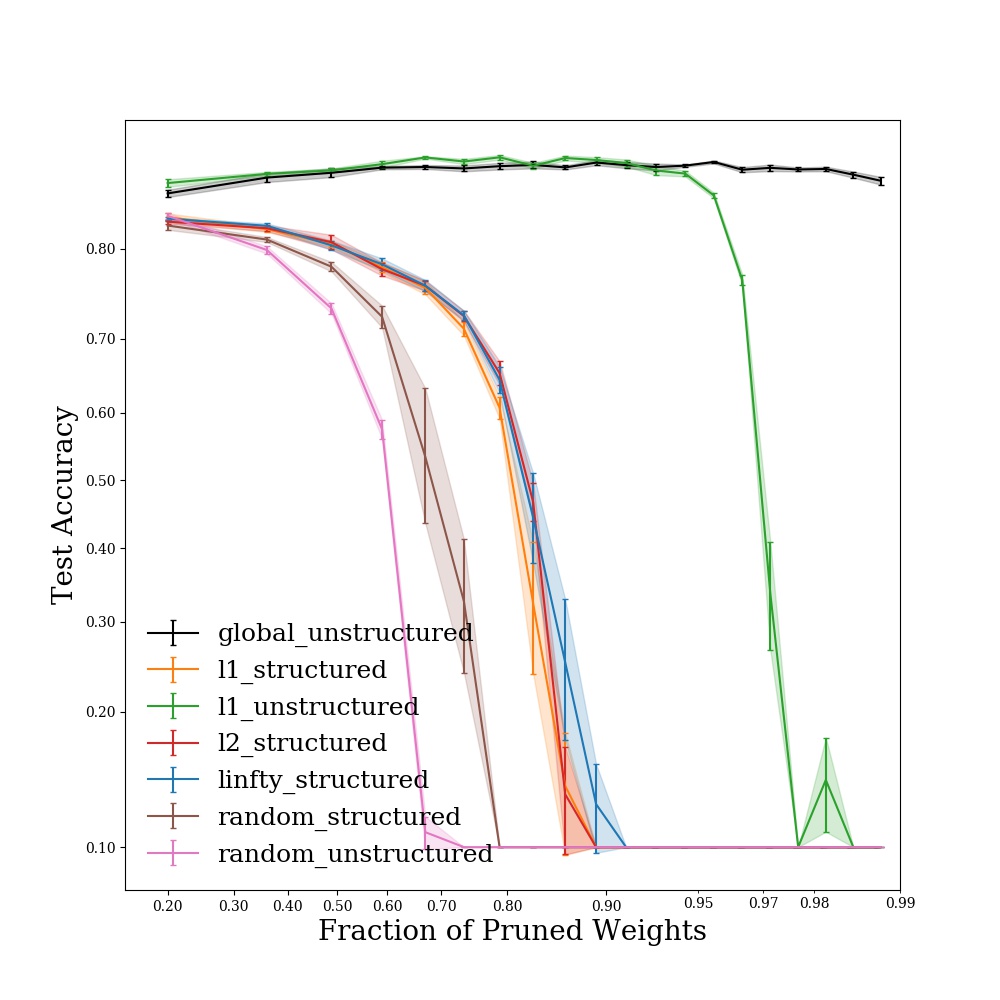}\label{fig:cifar10_alexnet}}
   \hspace{0.1em}
  \subfloat[AlexNet on MNIST]{\includegraphics[width=0.3\textwidth]{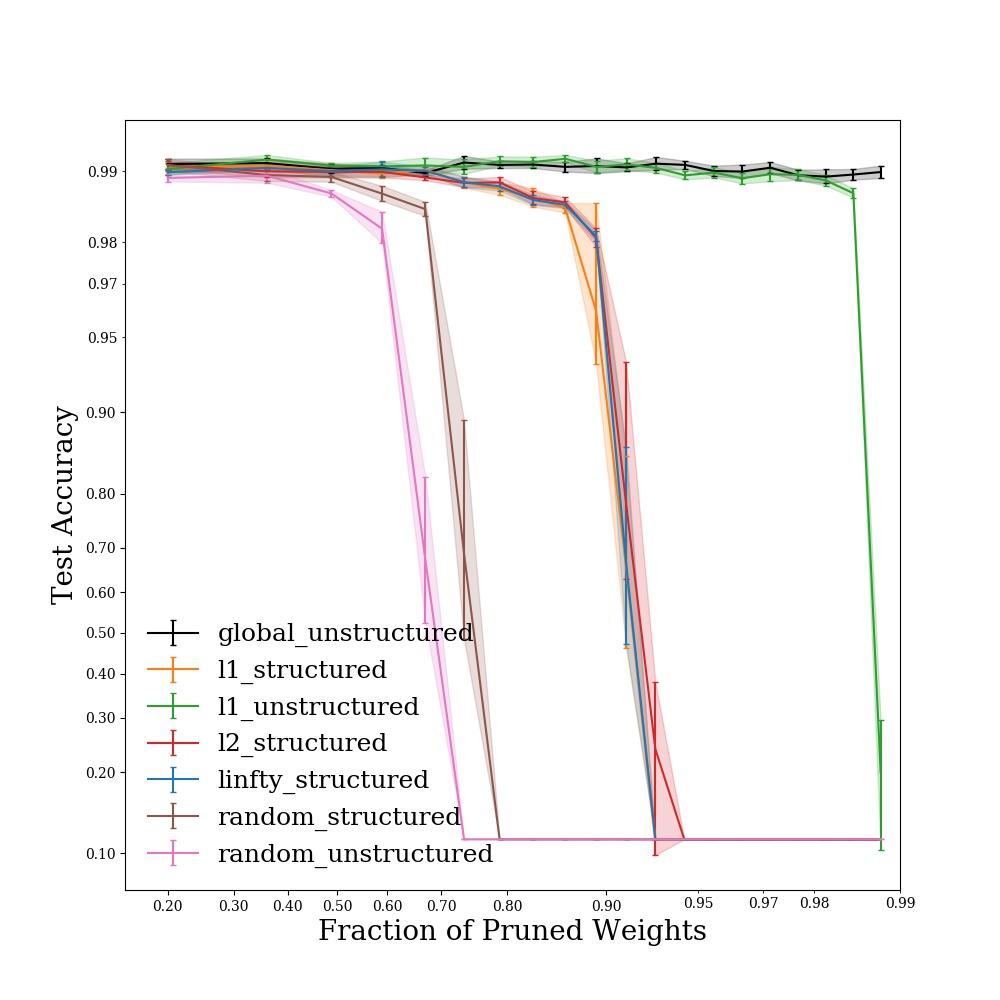}\label{fig:mnist_alexnet}}
   \hspace{0.1em}
  \subfloat[AlexNet on SVHN]{\includegraphics[width=0.3\textwidth]{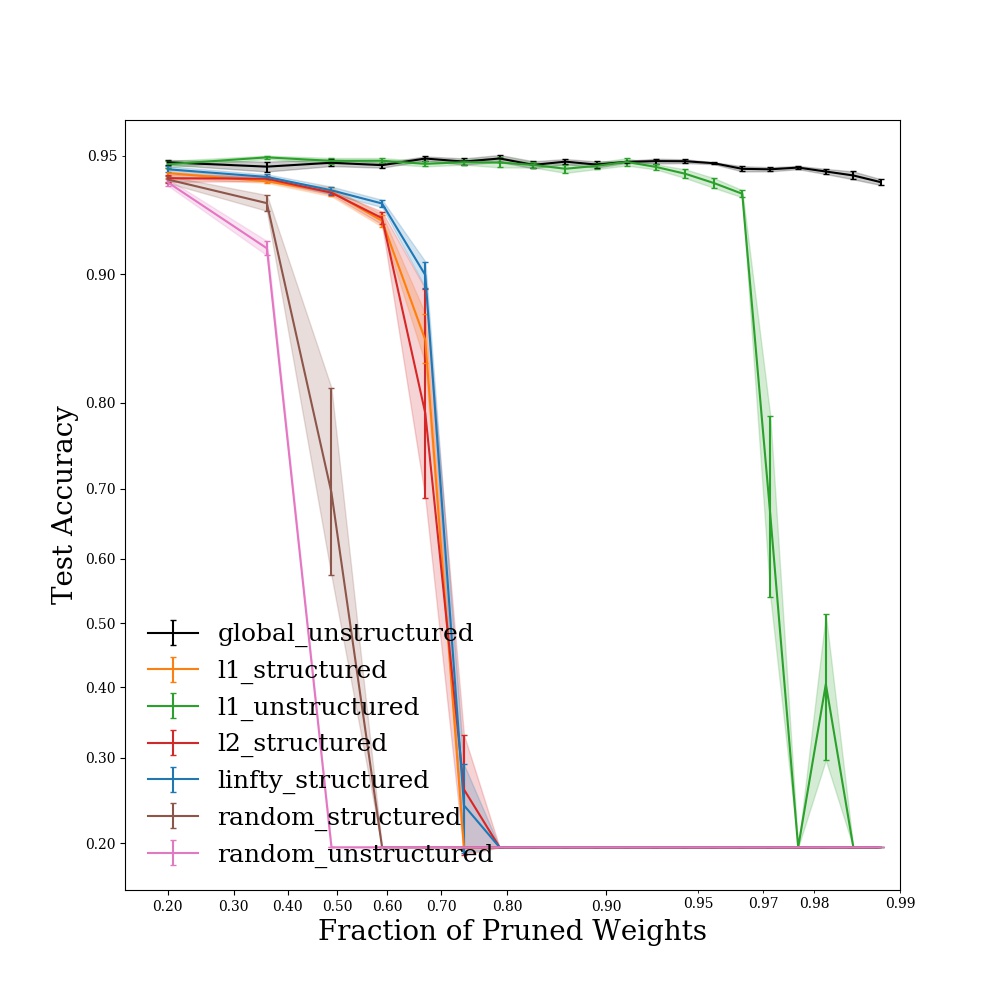}\label{fig:svhn_alexnet}}
  \\
  \subfloat[ResNet18 on CIFAR-100]{\includegraphics[width=0.28\textwidth]{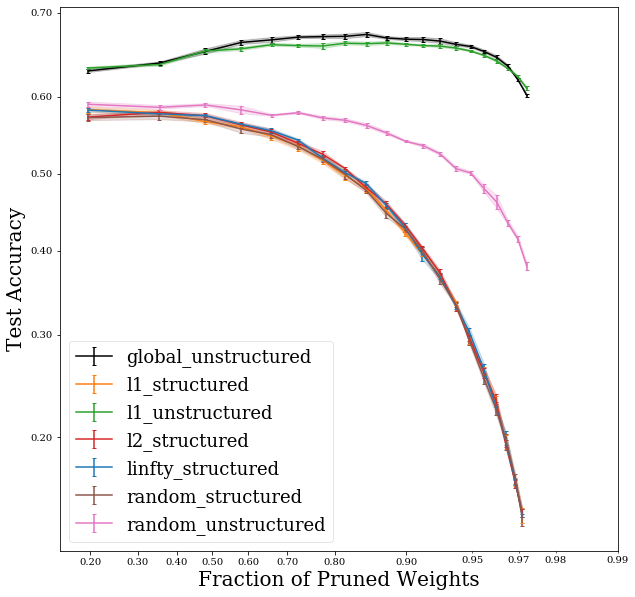}\label{fig:cifar100_resnet}}
  \hspace{0.1em}
  \subfloat[LeNet on CIFAR-100]{\includegraphics[width=0.3\textwidth]{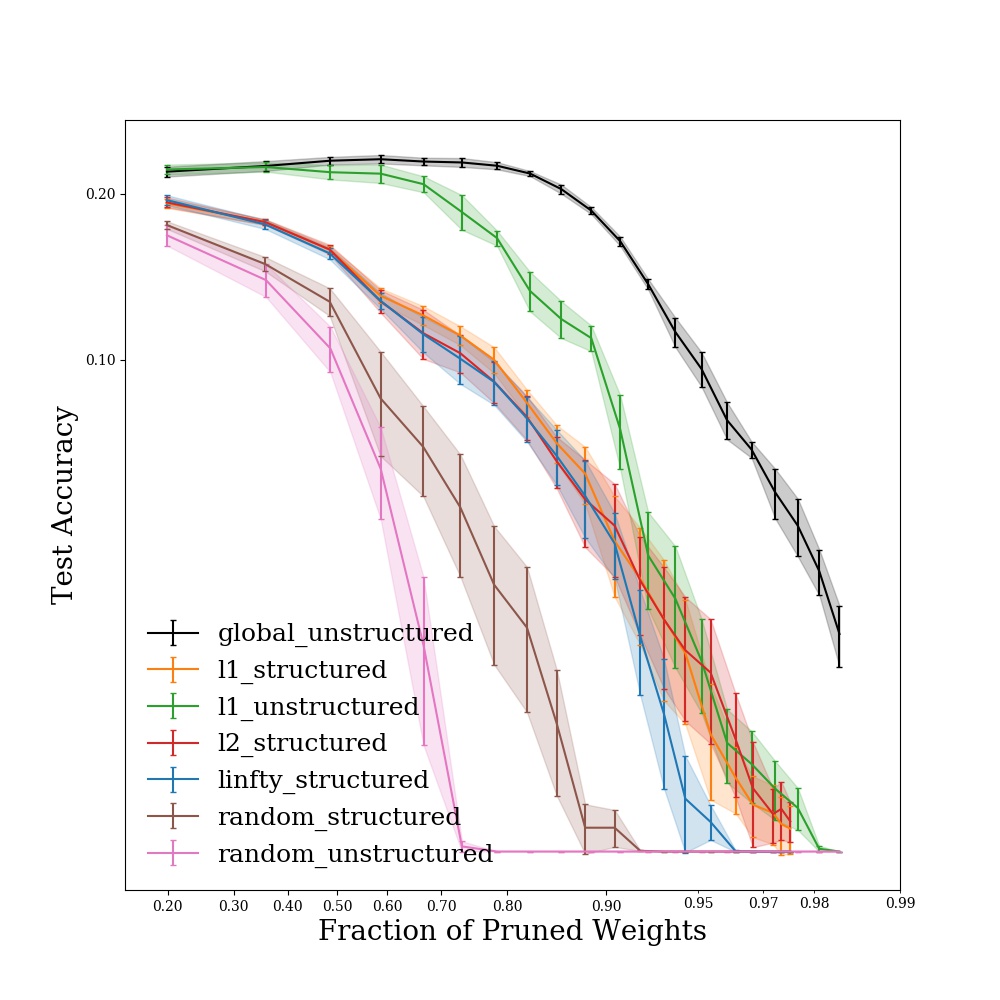}\label{fig:cifar100_lenet}}
  \hspace{0.1em}
  \subfloat[LeNet on CIFAR-10]{\includegraphics[width=0.3\textwidth]{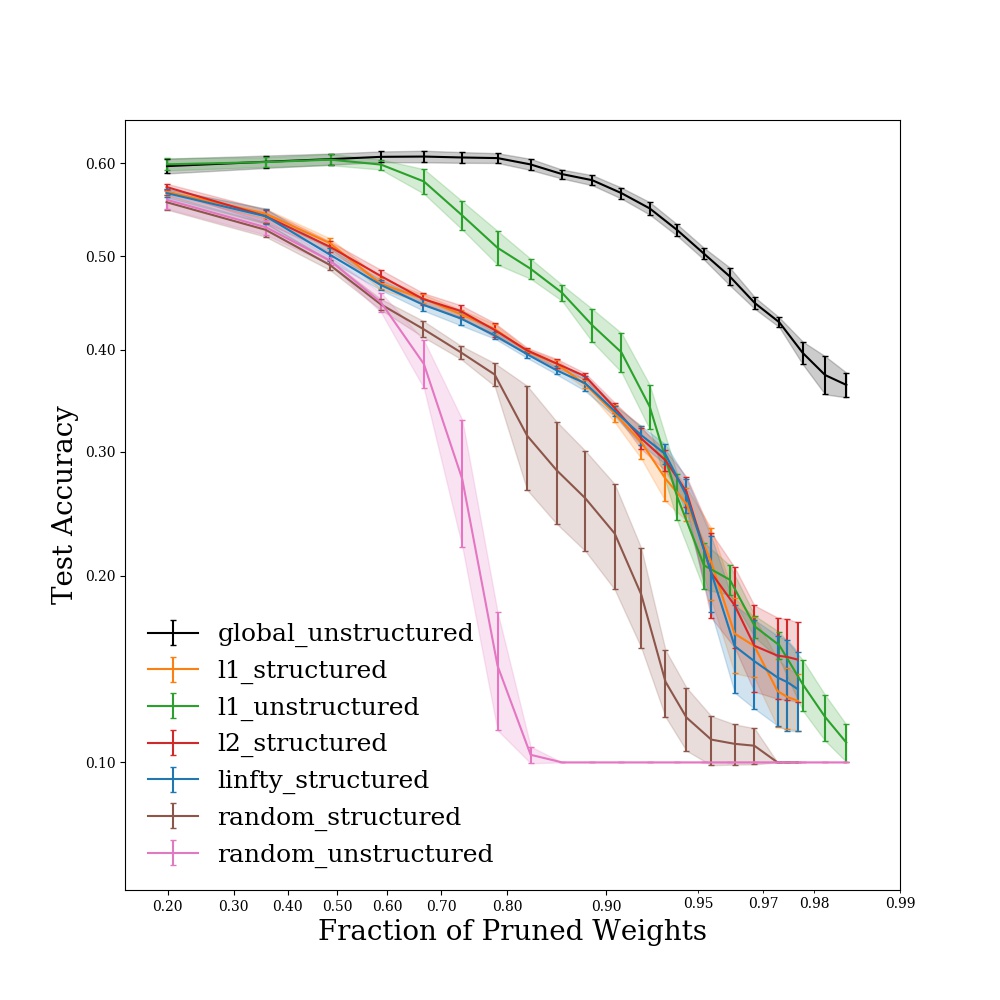}\label{fig:cifar10_lenet}} \\
\subfloat[LeNet on MNIST]{\includegraphics[width=0.3\textwidth]{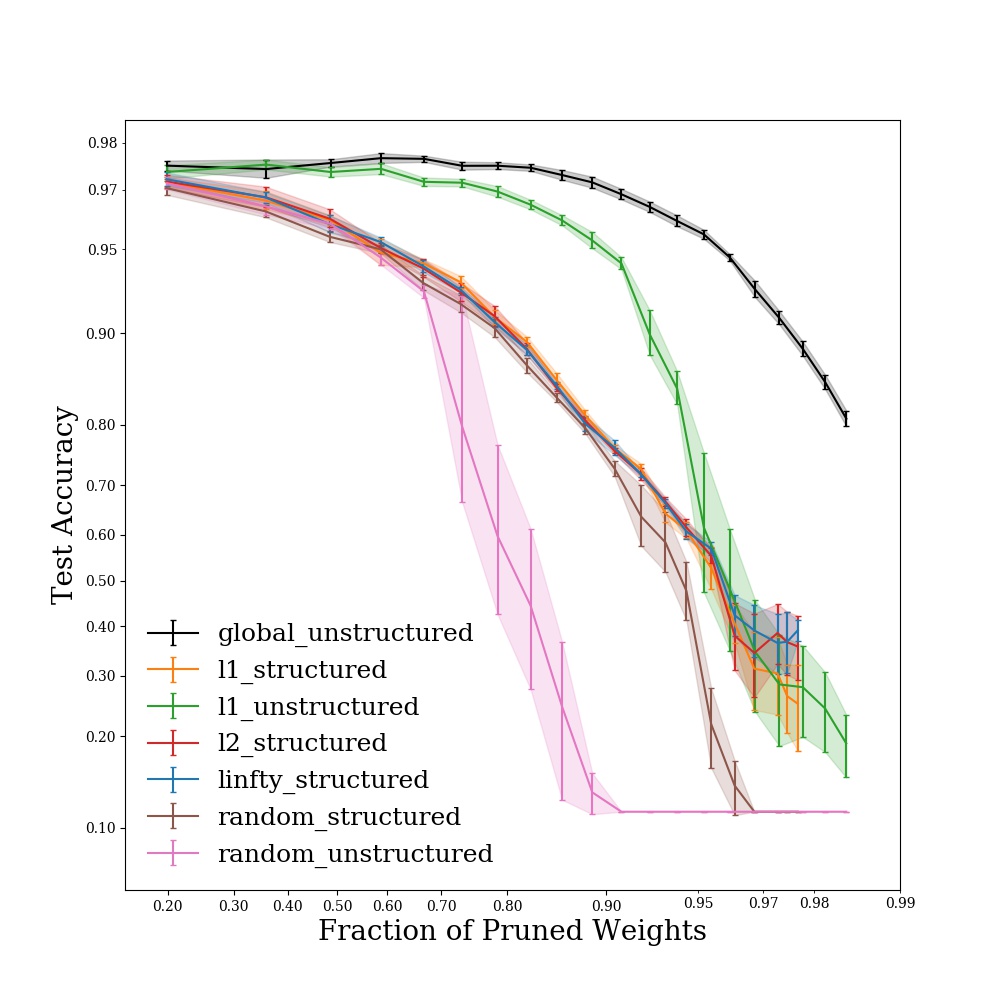}\label{fig:mnist_lenet}}
  \hspace{0.1em}
  \subfloat[LeNet on KMNIST]{\includegraphics[width=0.3\textwidth]{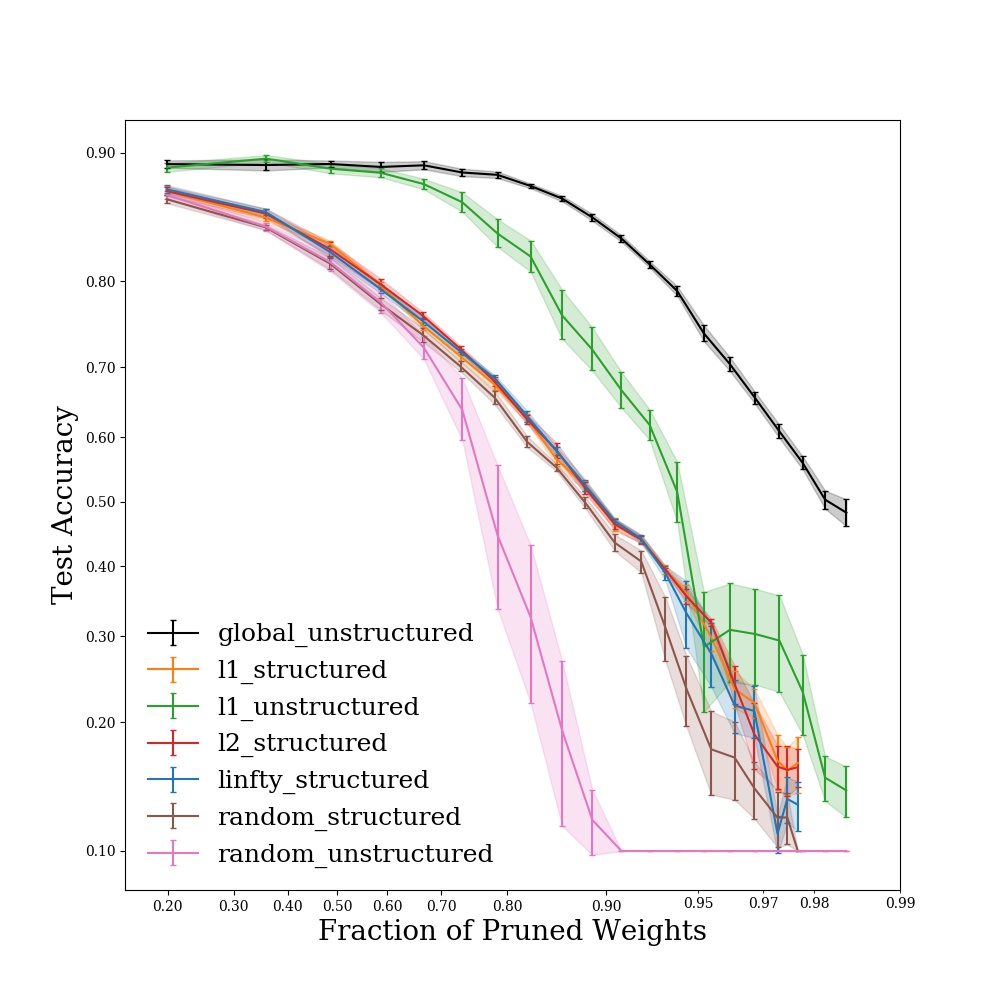}\label{fig:kmnist_lenet}}
  \hspace{0.1em}
  \subfloat[LeNet on SVHN]{\includegraphics[width=0.3\textwidth]{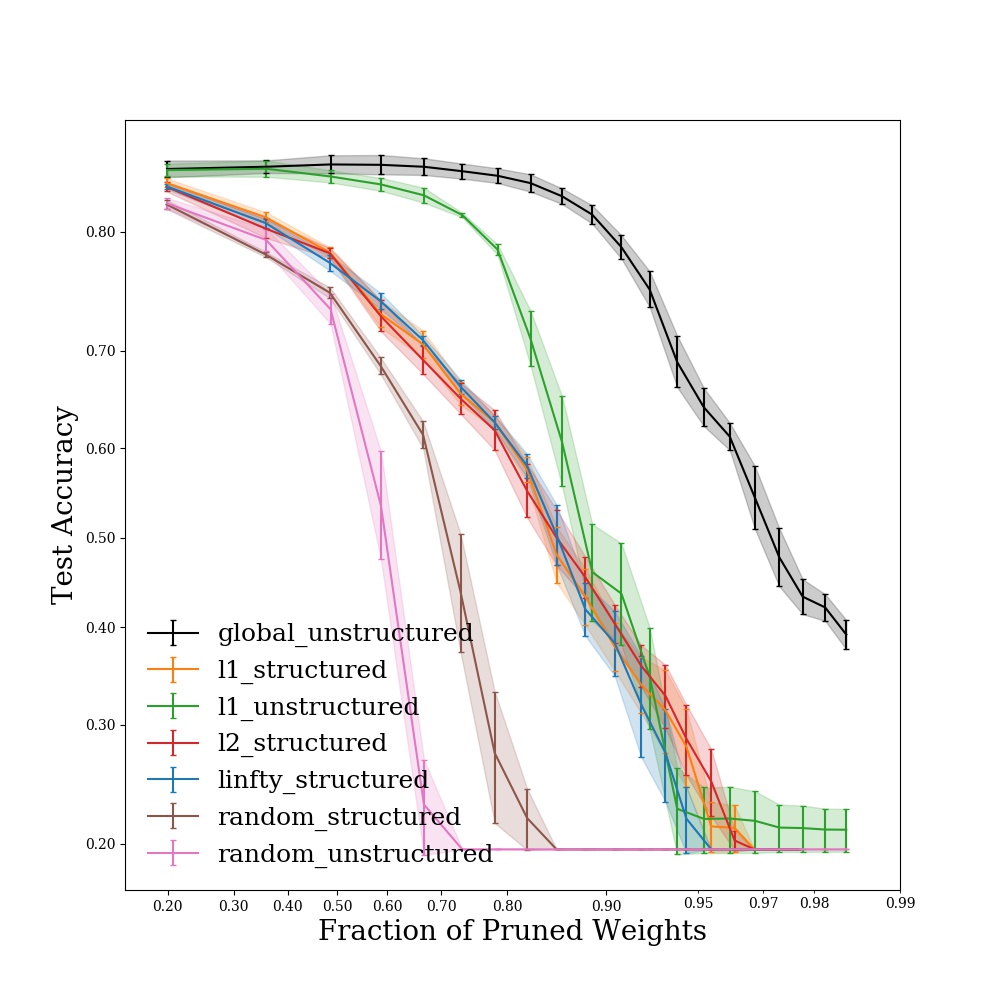}\label{fig:svhn_lenet}}
   \caption{Test accuracy for SGD-trained, sparsified models, pruned using seven different pruning techniques (\textcolor{red}{$L_2$-structured}, global unstructured, \textcolor{green}{$L_1$-unstructured}, \textcolor{orange}{$L_2$-structured} \textcolor{blue}{$L_\infty$-structured}, \textcolor{brown}{random structured}, and \textcolor{pink}{random unstructured}), and rewound to initial weight values after each pruning iteration.}\label{fig:accspars_more}
\end{figure}

Fig.~\ref{fig:accspars_more} shows additional accuracy-sparsity trade-off curves over various combinations of datasets and models. Both axes are plotted in logit scale. Similar curves for all dataset-model combinations are available upon request.

The power of global unstructured pruning, in black, lies in its flexibility to compare parameters across layers, although its advantage over its layer-wise counterpart decreases for more expressive and over-parametrized base architectures. It is also know that it is common to observe increases, as opposed to decreases, in total accuracy for moderate pruning fractions.

As a reminder, however, our simple training strategy, with constant learning rate and number of training epochs, employed to partially train the networks prior to pruning, is sub-optimal to achieve maximum performance; that, in fact, is not the goal of these training rounds, as the networks are not trained to full convergence or zero training loss. Therefore, it is not recommended to draw generalized conclusions about the maximum achievable performance of any lottery ticket found through any pruning technique by simply relying on the results in Fig.~\ref{fig:accspars_more}. To assess the full performance of these lottery tickets at any sparsity level, we suggest retraining them from scratch with an optimized training strategy and well sourced, case-specific hyper-parameters. In other words, the results shown here are intended to show the performance of these networks at the point in which the training was stopped prior to sparsifying the network further, and it is therefore not meant to be interpreted as these tickets' maximum achievable performance.

\section{Finetuning vs. reinitializing}
\label{sec:signbased_reinit}
\label{sec:appfinetuning}

The ``Lottery Ticket Hypothesis"~\cite{Frankle2018-po} postulates that rewinding weights to the initial (or early-stage~\cite{Frankle2019-vu}) values after pruning is key to identifying lucky sub-networks. In these experiments, we focus primarily on establishing the effect of the choice of weight handling strategy (finetuning vs. rewinding) on the structure of the masks of the resulting LTs.

\begin{figure}
\centering
  \subfloat{\includegraphics[width=0.32\textwidth]{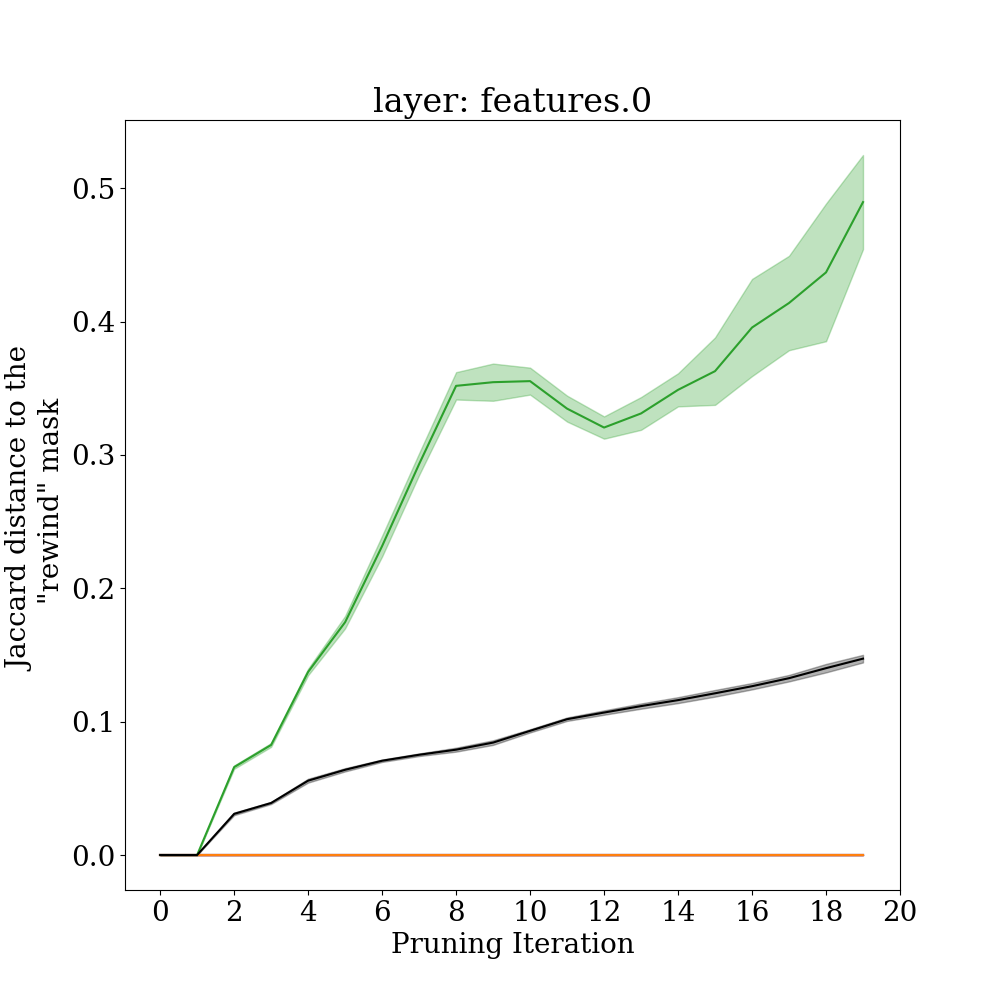}}
  \hspace{0.1em}
  \subfloat{\includegraphics[width=0.32\textwidth]{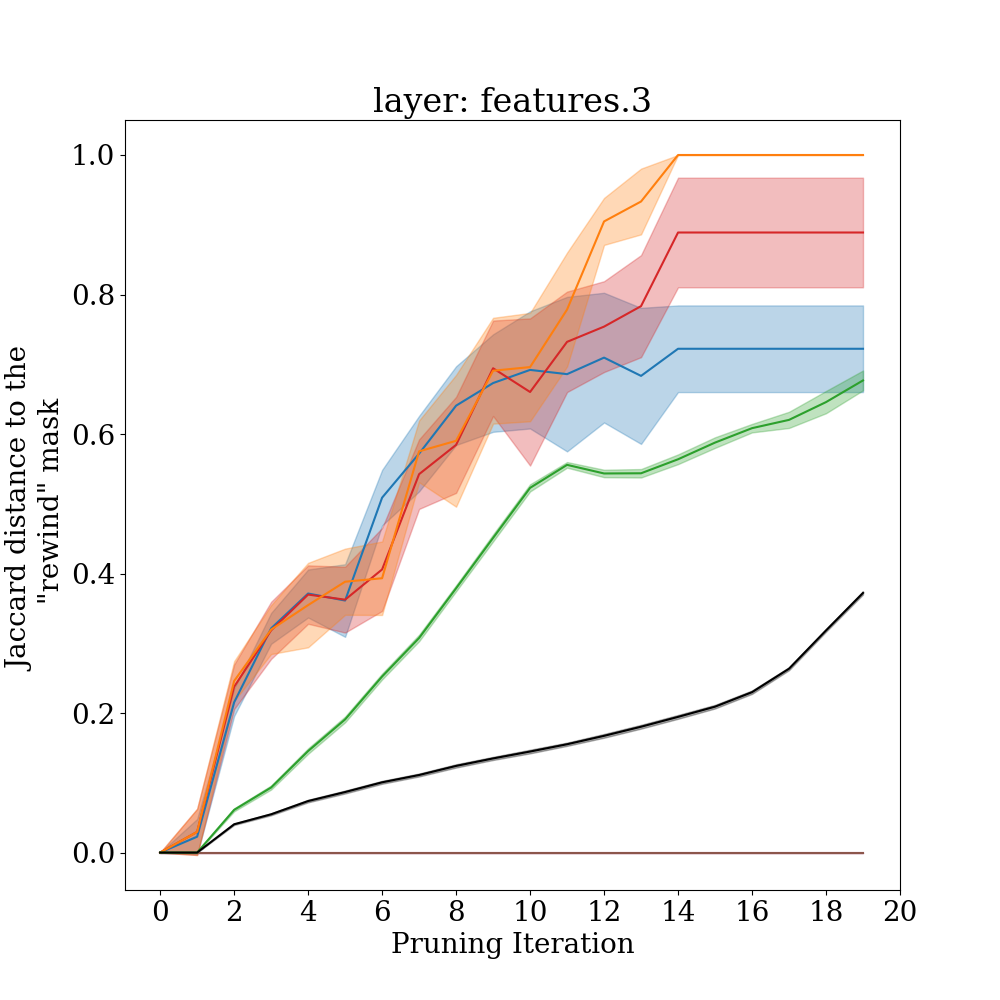}}
    \hspace{0.1em}\\
  \subfloat{\includegraphics[width=0.32\textwidth]{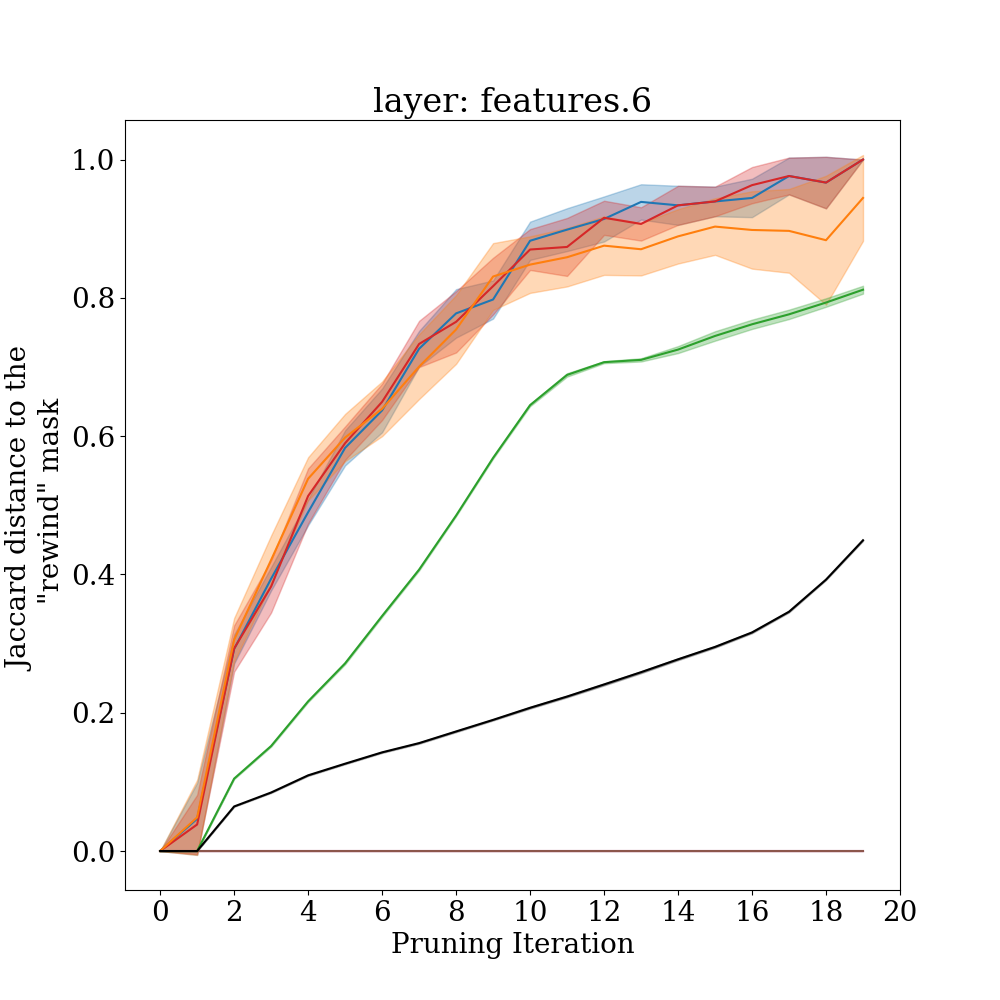}}
      \hspace{0.1em}
  \subfloat{\includegraphics[width=0.32\textwidth]{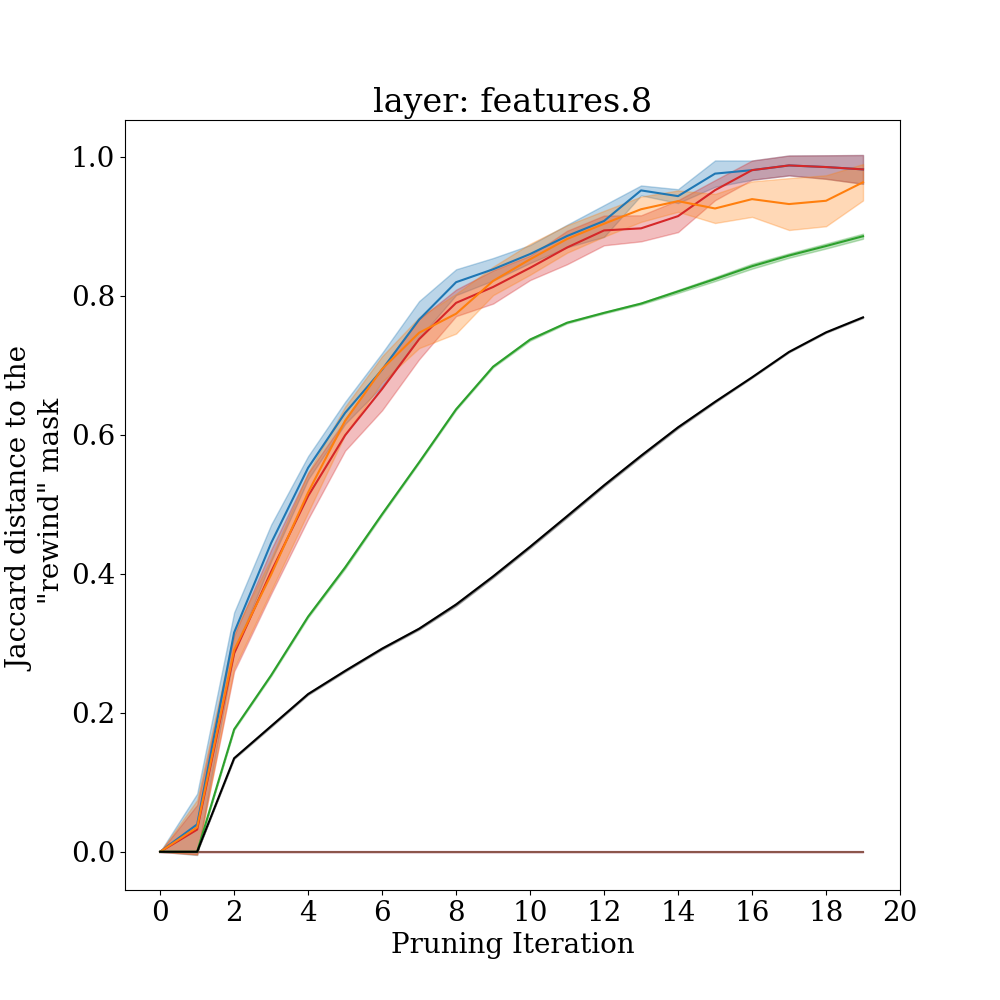}}\\
      \hspace{0.1em}\\
  \subfloat{\includegraphics[width=0.32\textwidth]{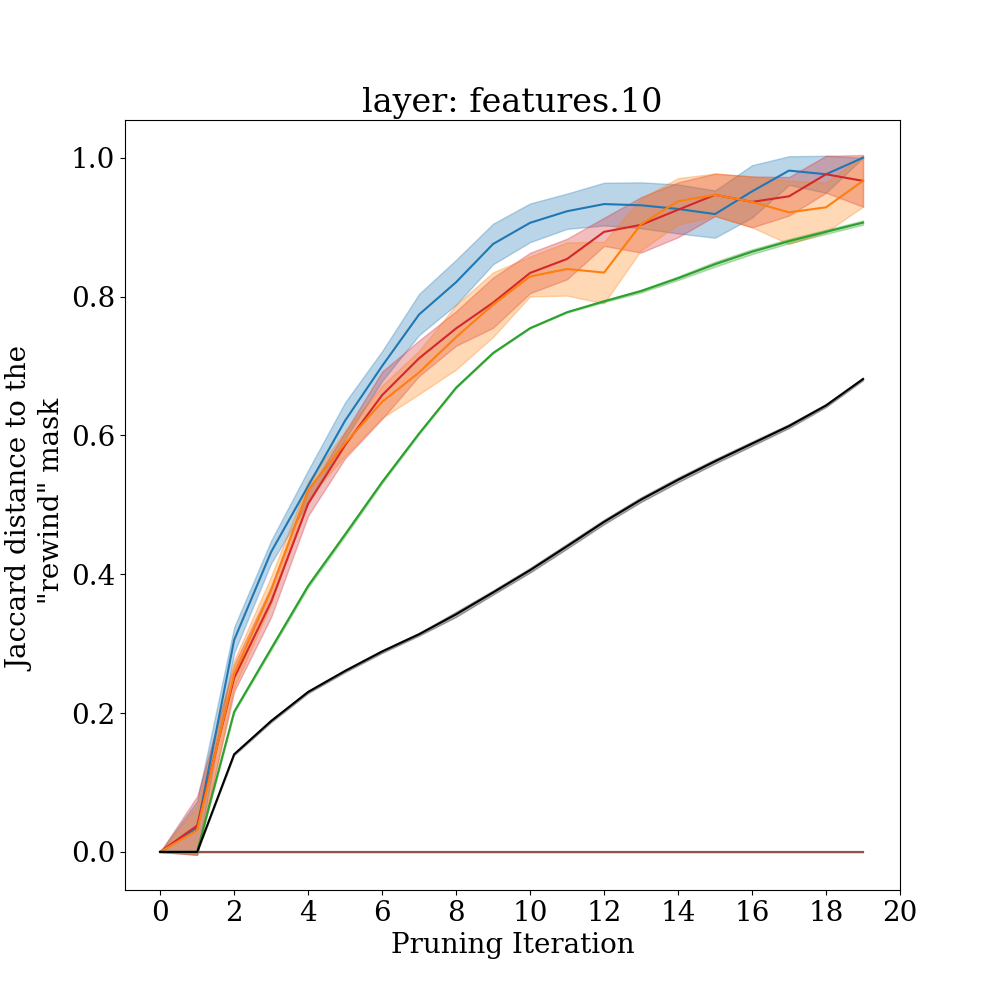}}
        \hspace{0.1em}
  \subfloat{\includegraphics[width=0.32\textwidth]{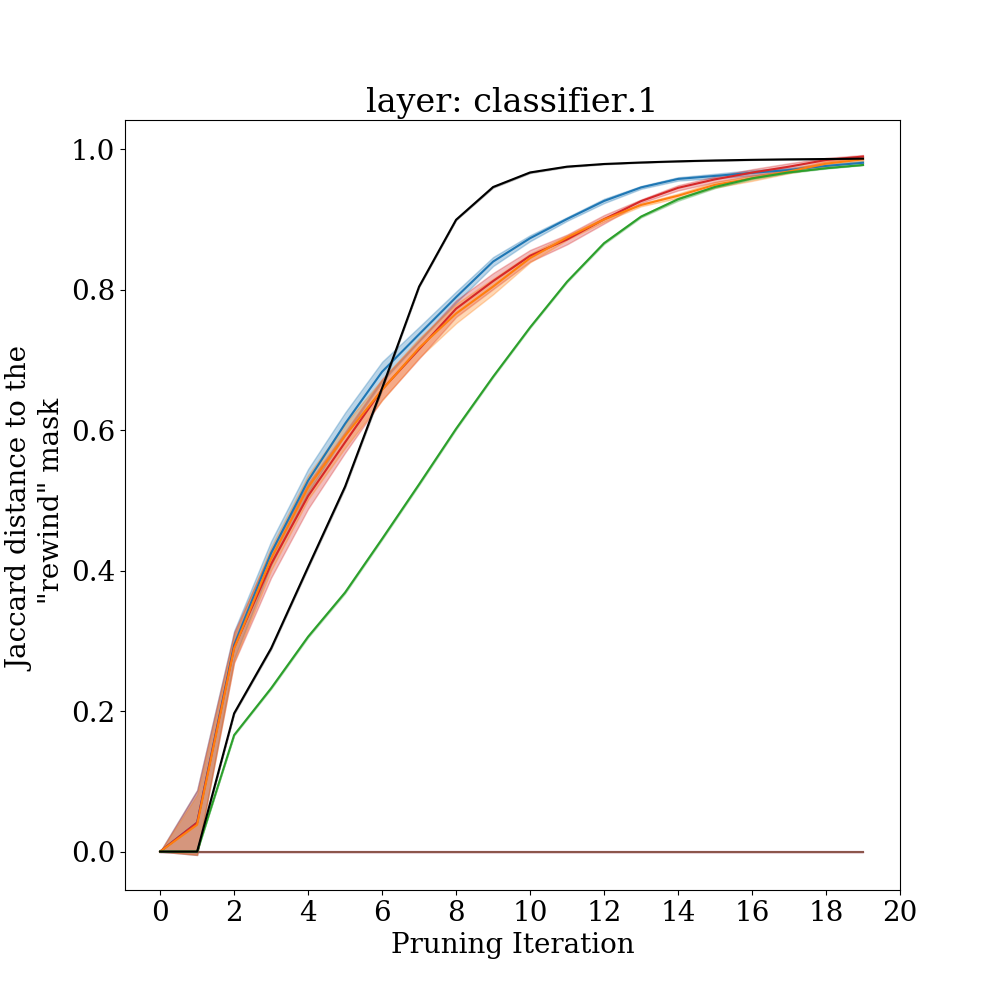}}
          \hspace{0.1em}\\
  \subfloat{\includegraphics[width=0.32\textwidth]{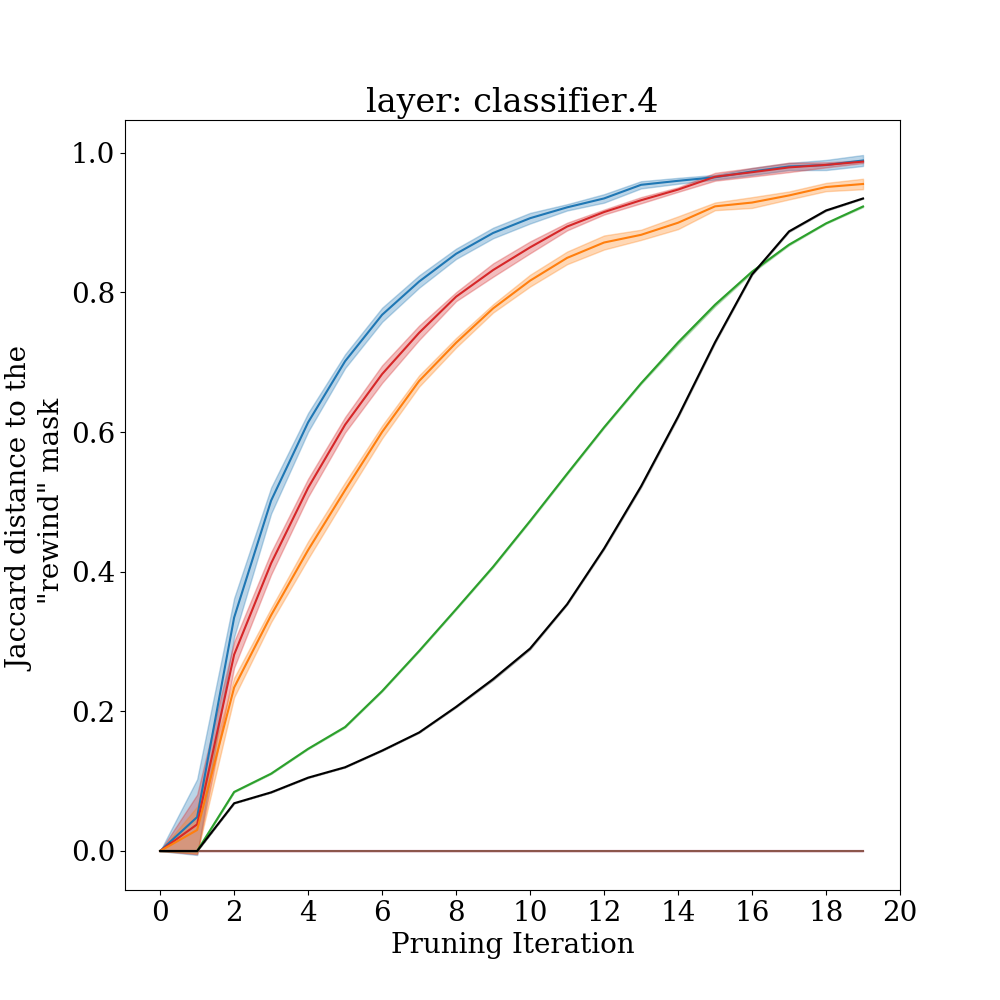}}
            \hspace{0.1em}
  \subfloat{\includegraphics[width=0.32\textwidth]{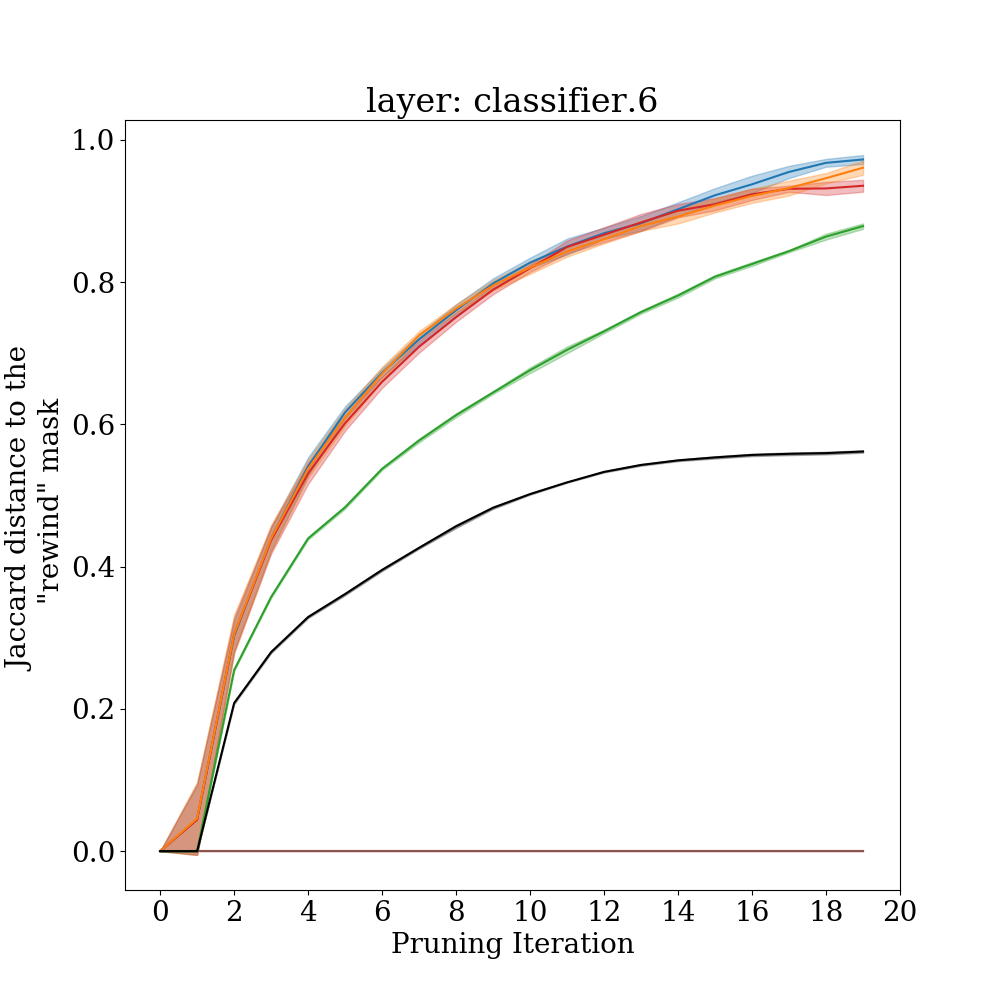}}
  \vspace{0.5cm}
   \caption{Layer-wise Jaccard distance between masks found by pruning AlexNet on MNIST and them rewind weights, and those found by finetuning after pruning, as a function of the pruning iteration, conditional on identical seed, across six seeds, and identical pruning technique, for \textcolor{red}{$L_2$-structured}, global unstructured, \textcolor{green}{$L_1$-unstructured}, \textcolor{orange}{$L_1$-structured}, \textcolor{blue}{$L_\infty$-structured}, \textcolor{brown}{random structured}, \textcolor{pink}{random unstructured} pruning.}\label{fig:jacc_finetune}
\end{figure}

The Jaccard distance, or any other similarity measure among masks (\textit{e.g.}, the Hamming distance), can be adopted to quantify the effects of the choice of finetuning or reinitializing weights after pruning. As expected, the nature of the connectivity structure that emerges in an iterative series of pruning and weight handling steps depends not only on the pruning choice but also on how weights are handled after pruning. Finetuning yields significantly different masks from rewinding, for all pruning techniques, although the difference is greater for magnitude-based structured pruning than for magnitude-based unstructured pruning (Fig.~\ref{fig:jacc_finetune}).
The difference (quantified in terms of the Jaccard distance) appears to grow logarithmically in the number of pruning iterations for magnitude-based structured pruning techniques; for magnitude-based unstructured pruning techniques, instead, the growth of the Jaccard distance exhibits more complex trends, with layer-by-layer differences.

\section{Effective pruning rate}
\label{sec:eff_pruning_rate}
A subtle implementation detail involves the way in which the fraction of pruned weights is computed, especially in convolutional layers pruned with structured pruning techniques. Take, for example, the architectures used in this work. When applying pruning to produce results like the ones displayed in Fig.~\ref{fig:accspars_more}, there are two ways to define the number of weights that get pruned (displayed along the x-axis). The definition used in the paper uses the fraction of weights \textit{explicitly} pruned by the decision rule that produces the mask for each layer. However, this does not account for the \textit{implicitly} pruned weights, \textit{i.e.} those weights that, due to downstream pruning in following layers, are now disconnected from the output of the neural network. For this reason, they receive no gradient and their value never changes from that at initialization. They can technically take on any value, including 0, without affecting the output of the neural network. In other words, they could effectively be pruned without any loss of performance. If we include these weights in the effective fraction of pruned weights, structured pruning begins to appear competitive, especially at high pruning fractions, because its effective sparsity naturally tends to be higher. This also suggests that since the effective pruning rate per iteration is higher than 20\% for structured pruning techniques, one could experiment with lowering it, to avoid aggressive pruning and consequent loss in performance.

The number of parameters corresponding to various fractions of pruned weights in the models investigated in this work is listed in Table~\ref{tab:params}.

\begin{table}
\centering
\caption{Number of parameters (in thousands) left in each model at each pruning fraction. Columns labeled as $s$ corresponds to the number of parameters that are not explicitly zero-ed out by the pruning mask in the network; $s_\mathrm{eff}$ corresponds to the number of non-disconnected parameters.}
\vspace{0.5cm}
\begin{tabular}{@{}ccccccccccccccc@{}}
\toprule
\begin{tabular}[c]{@{}c@{}}Pruning\\ Iteration\end{tabular} & \multicolumn{3}{c}{LeNet}                                          & \multicolumn{3}{c}{AlexNet}                                           & \multicolumn{3}{c}{VGG11}                                             & \multicolumn{3}{c}{ResNet18}  & \\                          
\midrule
& $s$  & $s_\mathrm{eff}$   && $s$  & $s_\mathrm{eff}$ && $s$ & $s_\mathrm{eff} $     &  & $s$         & $s_\mathrm{eff}$ \\ 
\cmidrule{2-3}  \cmidrule{5-6}  \cmidrule{8-9} \cmidrule{11-12}
 0 & 48 & 48 &&  45,637 & 45,637 &&  103,048 & 103,042 &&  8,981 & 8,981 \\
 1 & 38 & 38 &&  36,512 & 36,512 &&   82,440 & 82,430 &&  7,221 & 7,221 \\
 2 & 30 & 30 &&  29,211 & 29,211 &&   65,954 & 65,942 &&  5,813 & 5,813 \\
 3 & 24 & 24 &&  23,371 & 23,371 &&   52,765 & 52,718 &&  4,687 & 4,687 \\
 4 & 19 & 19 &&  18,698 & 18,692 &&   42,214 & 42,126 &&  3,786 & 3,786 \\
 5 & 15 & 15 &&  14,960 & 14,945 &&   33,774 & 33,610 &&  3,065 & 3,065 \\
 6 & 12 & 12 &&  11,970 & 11,937 &&   27,021 & 26,800 &&  2,488 & 2,488 \\
 7 & 10 & 10 &&  9,578 & 9,505 &&   21,619 & 21,357 &&  2,027 & 2,027 \\
 8 & 8.3 & 8.3 &&  7,664 & 7,549 &&   17,297 & 17,005 &&  1,658 & 1,658 \\
 9 & 6.7 & 6.6&&  6,133 & 5,959 &&   13,840 & 13,549 &&  1,362 & 1,362 \\
 10 & 5.4 & 5.4&&  4,908 & 4,649 &&   11,074 & 10,792 &&  1,126 & 1,126 \\
 11 & 4.3 & 4.3&&  3,928 & 3,607 &&   8,861 & 8,608 &&  937 & 937 \\
 12 & 3.5 & 3.5&&  3,144 & 2,810 &&   7,091 & 6,873 &&  786 & 786 \\
 13 & 2.9 & 2.8&&  2,517 & 2,199 &&   5,675 & 5,492 &&  665 & 665 \\
 14 & 2.3 & 2.3&&  2,016 & 1,745 &&   4,542 & 4,399 &&  568 & 567 \\
 15 & 1.9 & 1.9&&  1,614 & 1,401 &&   3,636 & 3,523 &&  491 & 490 \\
 16 & 1.6 & 1.5&&  1,293 & 1,142 &&   2,911 & 2,825 &&  429 & 427 \\
 17 & 1.3 & 1.2&&  1,036 & 950 &&   2,331 & 2,260 &&  379 & 377 \\
 18 & 1.1 & 1.0 &&  831 & 812 &&   1,867 & 1,807 &&  340 & 337 \\
 19 & 0.9 & 0.8&&  666 & 663 &&   1,495 & 1,443 &&  308 & 303 \\
\bottomrule
\end{tabular}
\label{tab:params}
\end{table}

\section{Lottery Tickets from different pruning techniques}
\begin{figure}
\centering
  \subfloat{\includegraphics[width=0.25\textwidth]{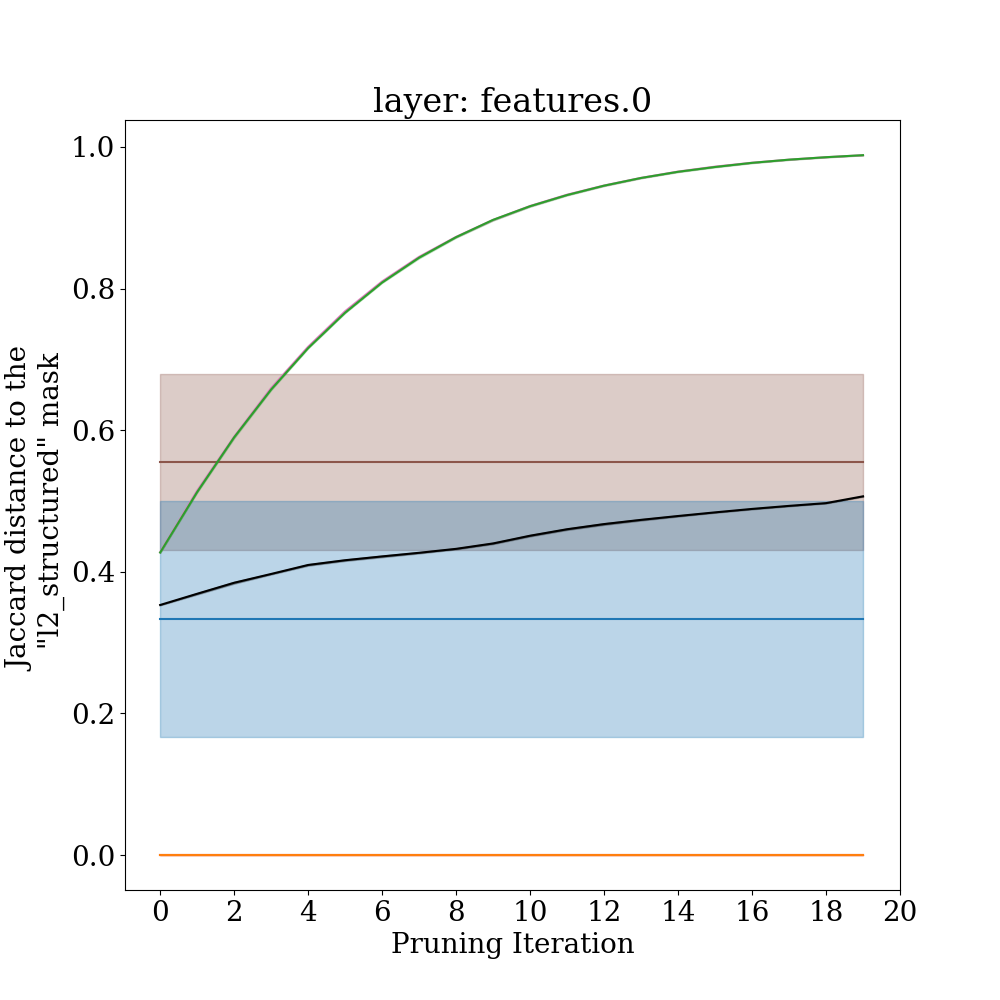}}
  \subfloat{\includegraphics[width=0.25\textwidth]{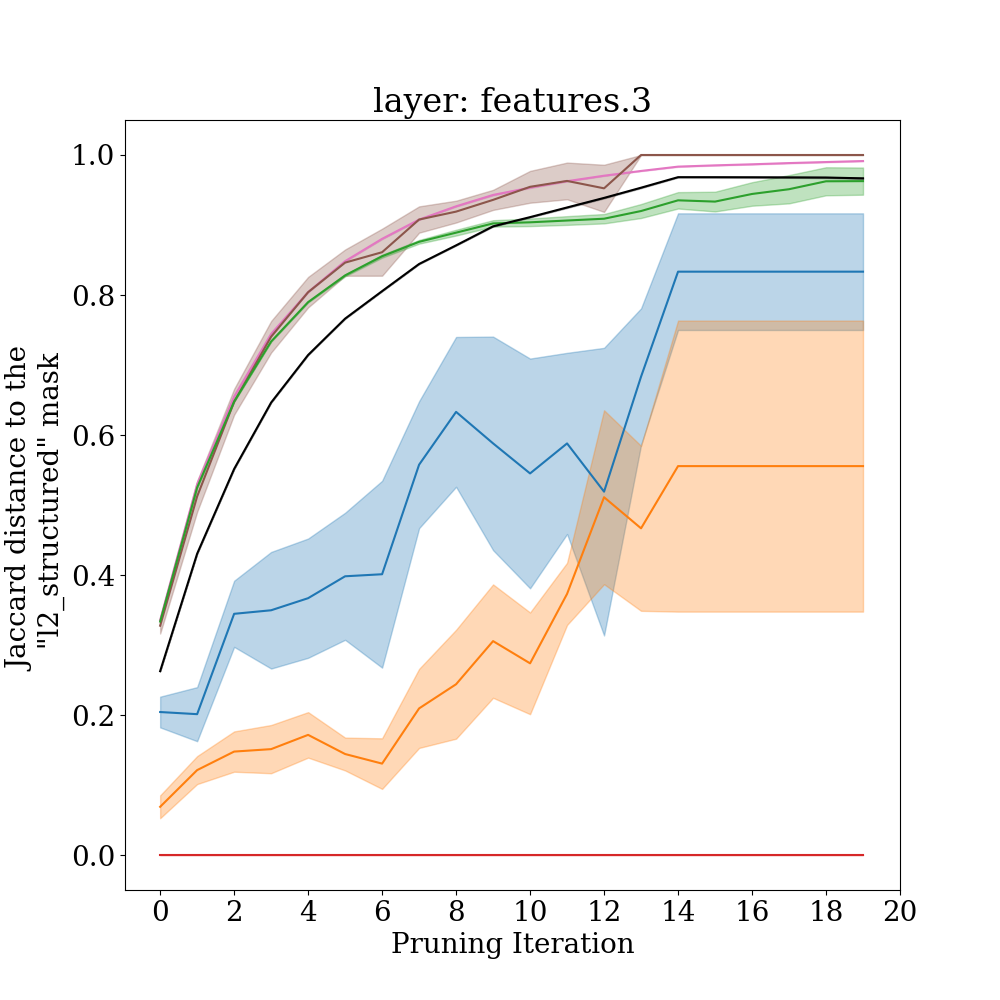}}
  \subfloat{\includegraphics[width=0.25\textwidth]{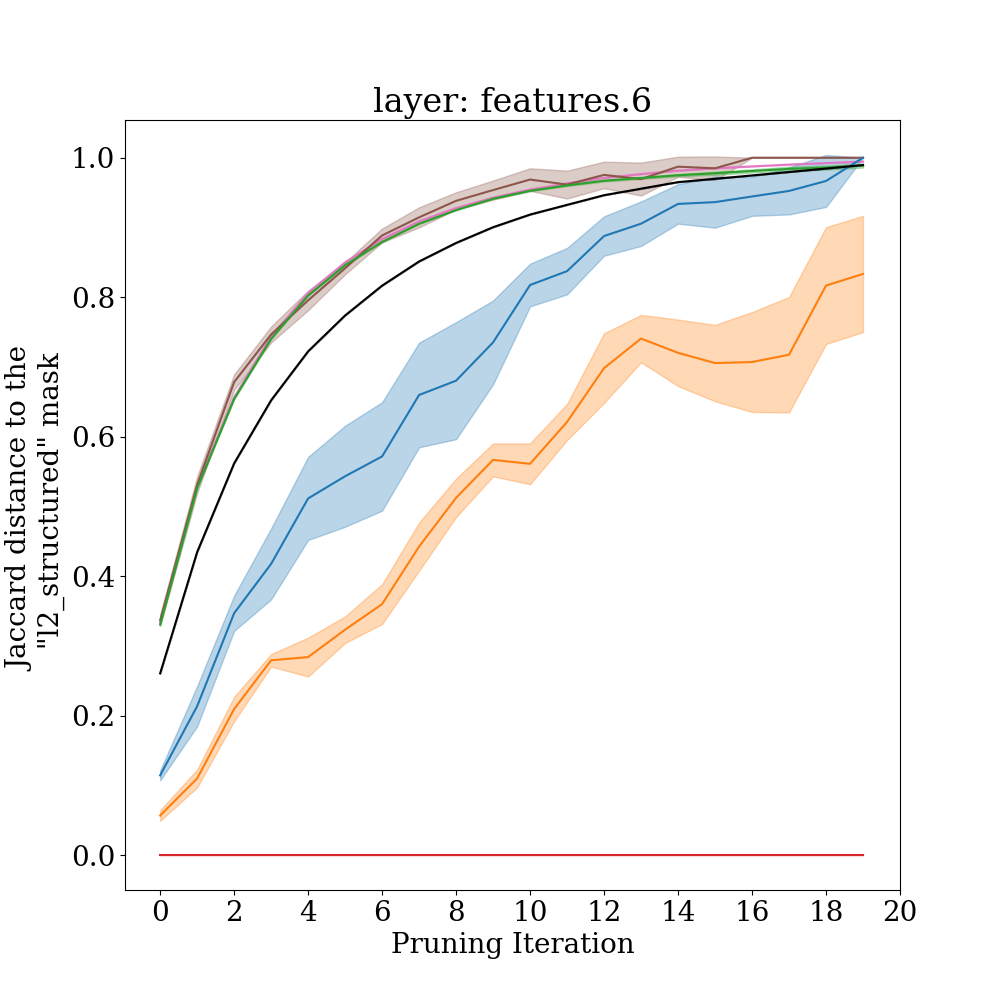}}
  \subfloat{\includegraphics[width=0.25\textwidth]{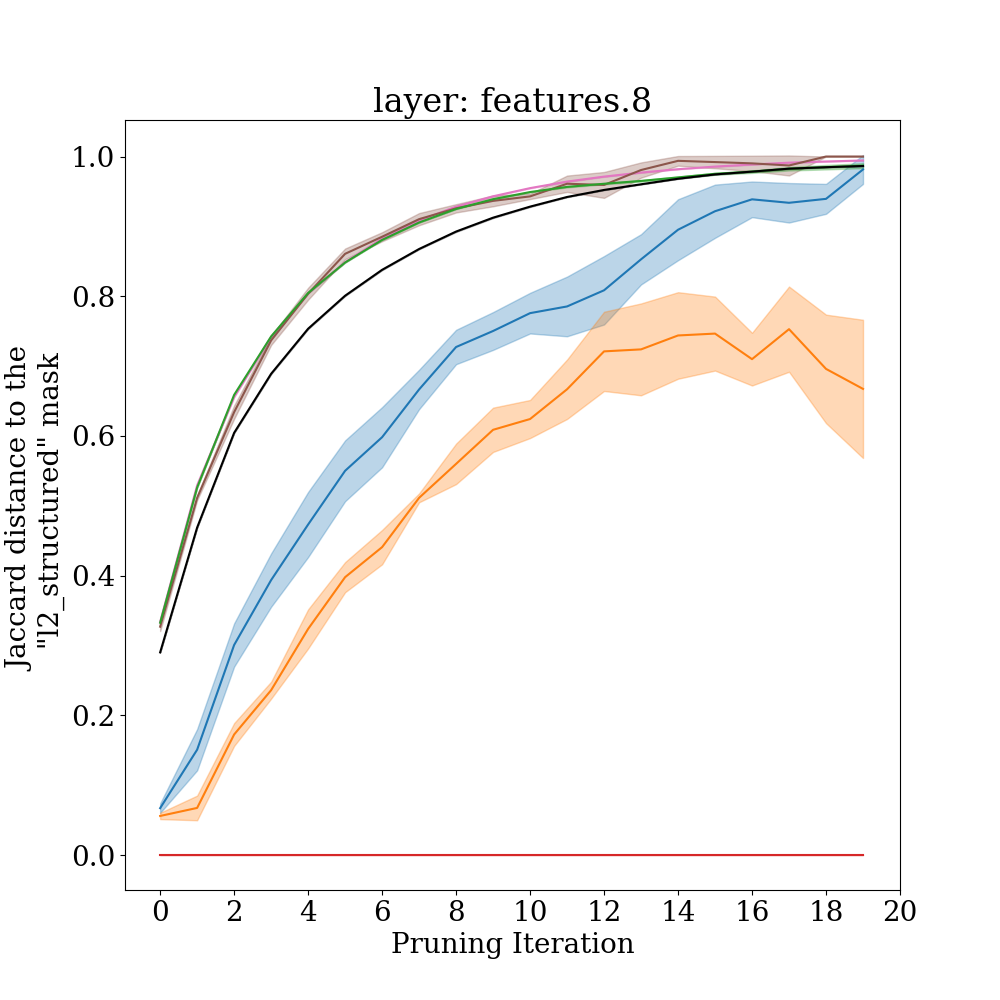}}\\
  \subfloat{\includegraphics[width=0.25\textwidth]{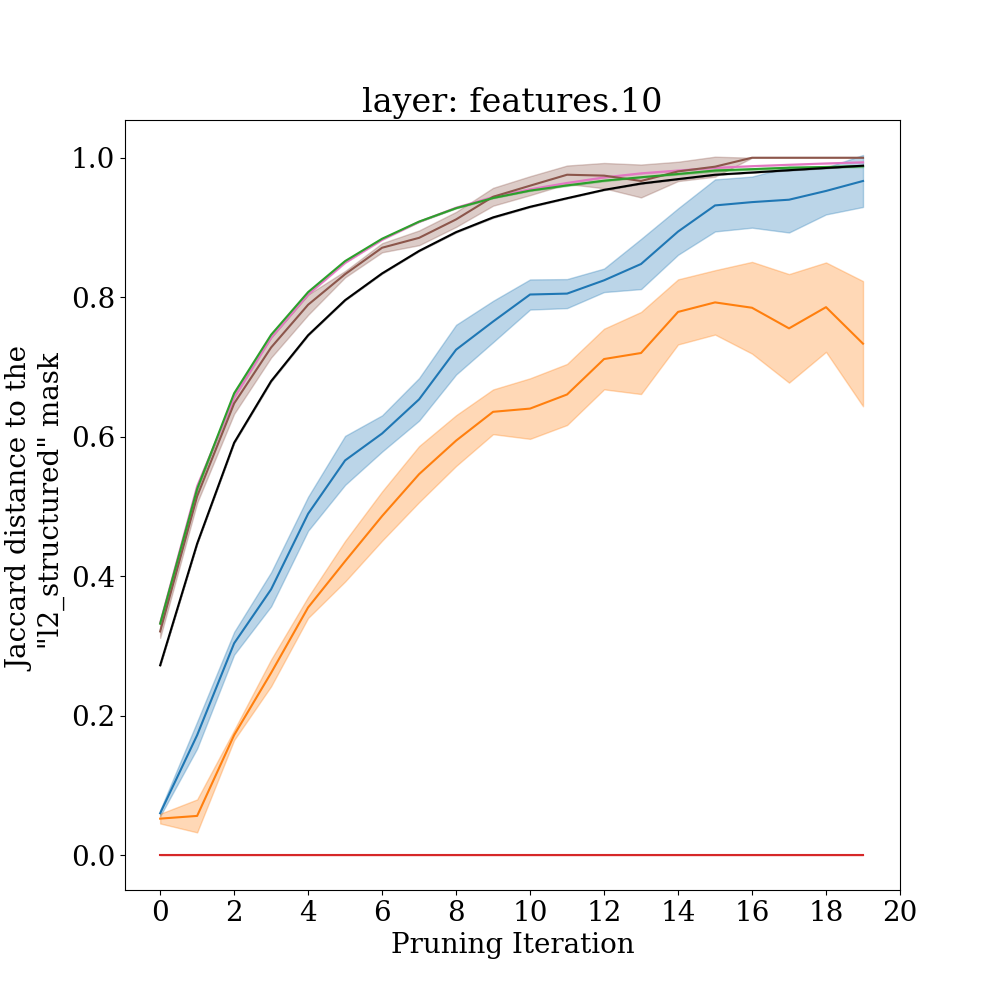}}
  \subfloat{\includegraphics[width=0.25\textwidth]{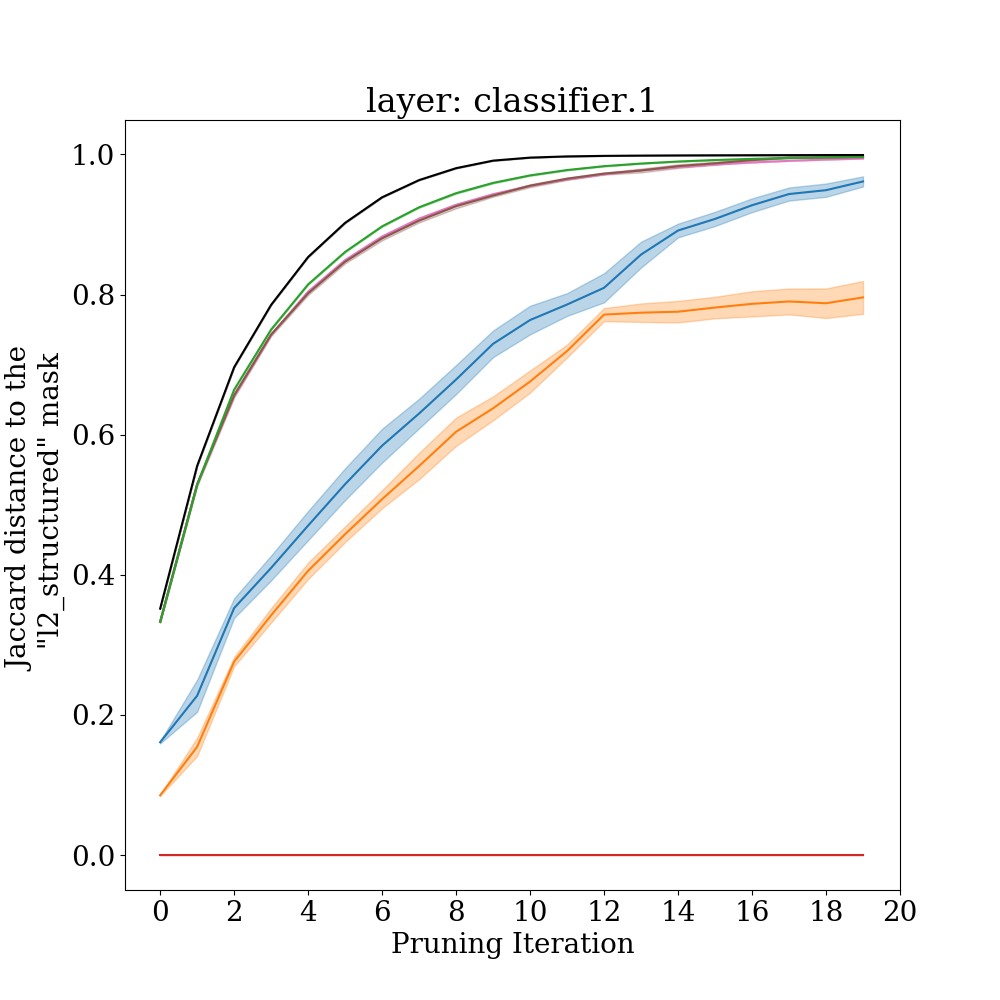}}
  \subfloat{\includegraphics[width=0.25\textwidth]{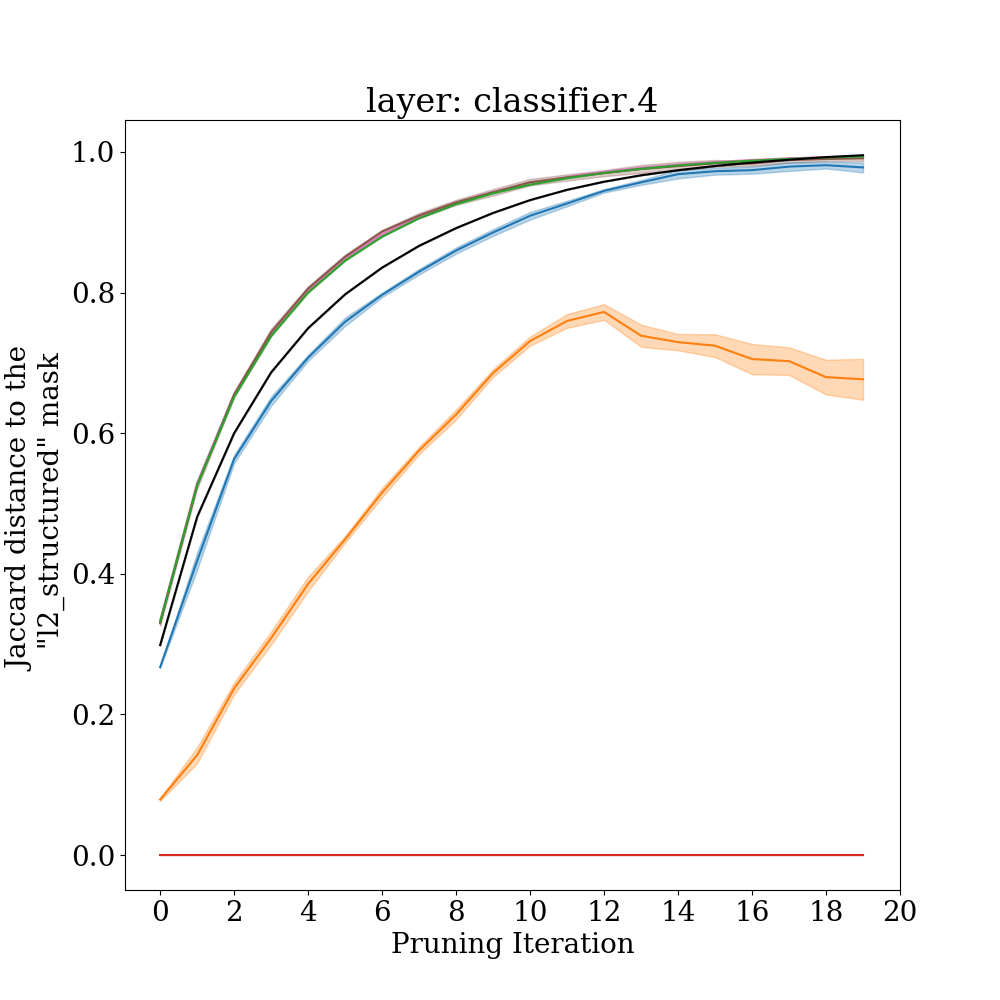}}
  \subfloat{\includegraphics[width=0.25\textwidth]{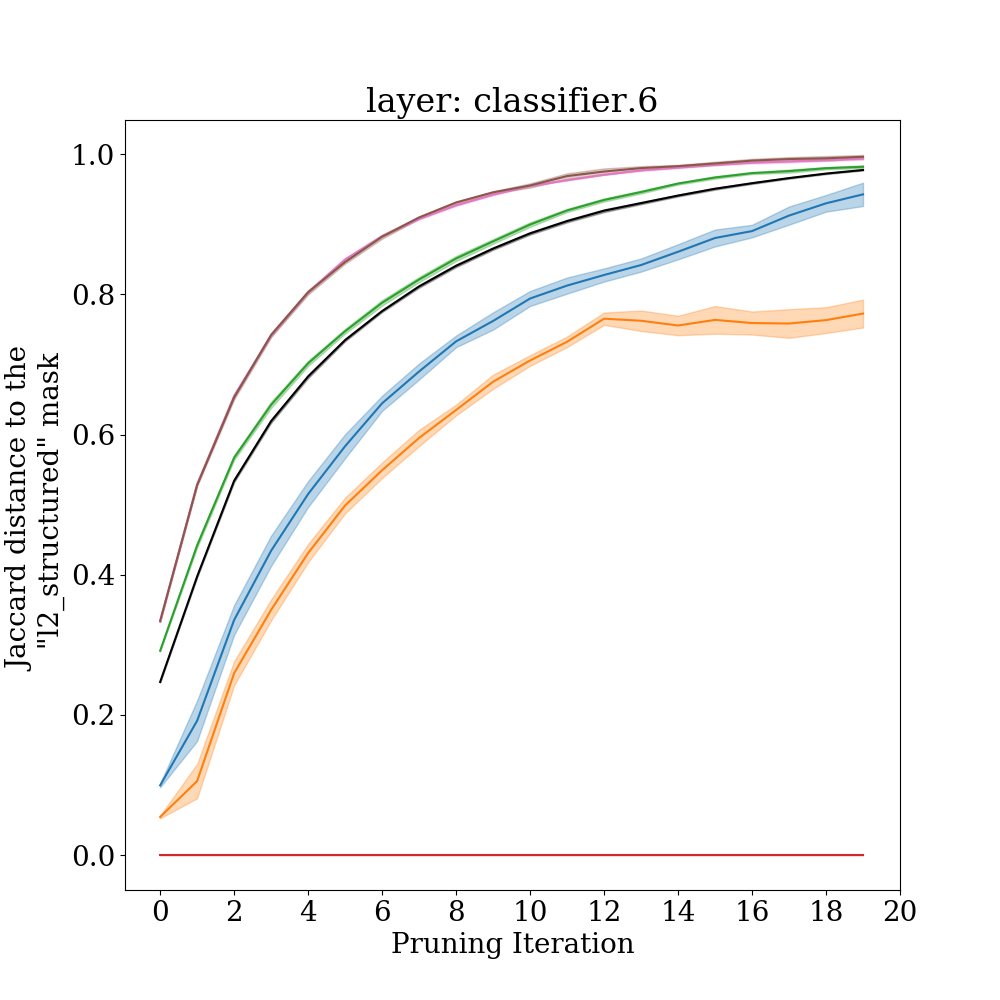}}
   \caption{Layer-wise Jaccard distance between masks found by pruning AlexNet on MNIST using \textcolor{red}{$L_2$-structured} pruning, and masks found by other pruning methods (global unstructured, \textcolor{green}{$L_1$-unstructured}, \textcolor{orange}{$L_1$-structured}, \textcolor{blue}{$L_\infty$-structured}, \textcolor{brown}{random structured}, \textcolor{pink}{random unstructured}), conditioned on identical seed, across six seeds. $L_1$-structured pruning yields the most similar masks to $L_2$-structured pruning, as expected.}\label{fig:jaccard_alexnet_mnist}
\end{figure}

Fig.~\ref{fig:jaccard_alexnet_mnist} shows an additional set of plots of the average Jaccard distance between the masks of lottery tickets found by different pruning techniques, this time for AlexNet trained on MNIST.

The structure of the masks generated by different pruning techniques begins to diverge rapidly in the early phases of pruning, and approaches a total Jaccard distance of 1 (no intersection over masks) as sparsity is increased towards 100\%. All distances are measured with respect to the mask generated by $L_2$ structured pruning, which is used as a baseline. For pairwise distances among masks over training iterations, please refer to Fig.~\ref{fig:jaccard_heats_prun}.

\subsection{Complementarity of Learned Solutions}
\label{ssec:heatmaps}

\begin{figure}
    \centering
	\subfloat[$19^\mathrm{th}$ pruning iteration]{
        \centering
        \includegraphics[width=0.45\textwidth]{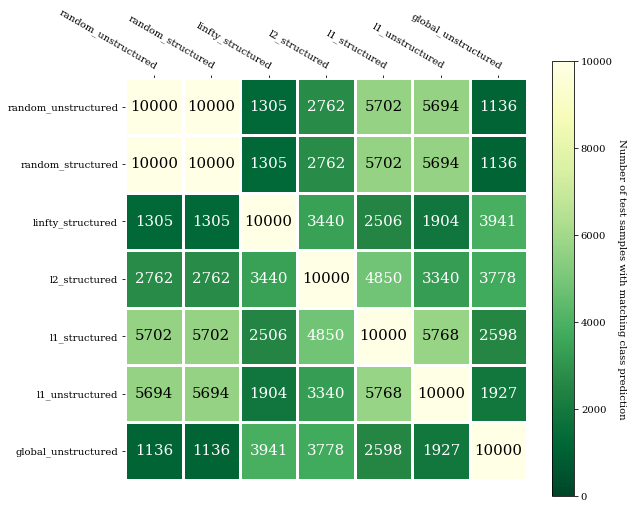}
    } 
    \caption{Number of examples in the MNIST test set over which the sub-networks obtained through each pruning technique agree on the prediction, on average (over 6 experimental seeds).}
    \label{fig:heatmaps}
\end{figure}

The dissimilarity of solutions can be explored by looking at the heat maps of the agreement in average class prediction across LTs obtained through different pruning techniques. Fig.~\ref{fig:heatmaps} provides these visualizations for the $19^\mathrm{th}$ pruning iteration. 

\section{Generating prêt-à-porter lottery tickets}
\label{appendix:gen-pap}
The prêt-à-porter tickets considered in this work were sourced from low sparsity bespoke tickets obtained on five 10-class classification tasks. Specifically, we used the bespoke tickets obtained via global unstructured pruning with rewinding, after only 2 regular pruning iterations. For LeNet tickets, for example, this corresponded to a bespoke ticket of 35.9\% sparsity. The intersection of the masks of these five bespoke tickets yielded a mask $\sM$ of sparsity ${\sim}{85\%}$, which is comparable to the sparsity of a bespoke ticket after 9 successive pruning iterations. By sourcing the five base tickets in parallel after only two pruning iterations, we effectively save the elapsed real time equivalent of 7 additional, sequential pruning iterations, without, however, any advantage in CPU time.

If a higher sparsity prêt-à-porter ticket was needed, one could have taken the intersection of bespoke ticket masks from a later pruning iteration.

\begin{figure}
\centering
  \subfloat[Pruning iteration 1]{\includegraphics[width=0.4\textwidth]{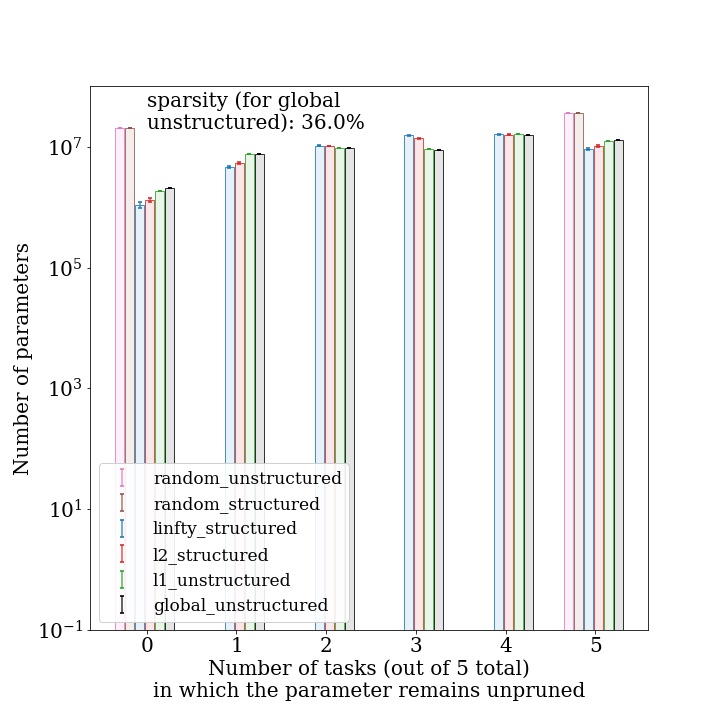}}
  \hspace{0.3em}
  \subfloat[Pruning iteration 6]{\includegraphics[width=0.4\textwidth]{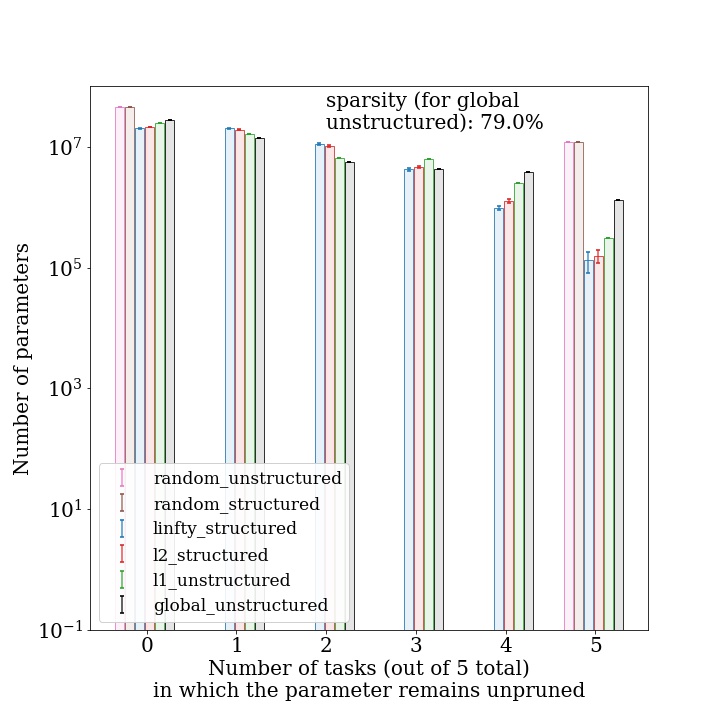}}
   \caption{Number of AlexNet weights that remain unpruned at each pruning iteration across 0 through 5 tasks (MNIST, FashionMNIST, KMNIST, CIFAR-10, SVHN) for each pruning technique: \textcolor{red}{$L_2$-structured}, global unstructured, \textcolor{green}{$L_1$-unstructured},  \textcolor{blue}{$L_\infty$-structured}, \textcolor{brown}{random structured}, and \textcolor{pink}{random unstructured}.}\label{fig:intersection_size_alex}
\end{figure}

Fig.~\ref{fig:intersection_size_alex} shows the number of parameters that remain unpruned in 0 through 5 of the chosen tasks for different kinds of pruning techniques, and across pruning iterations. Different color bars represent networks pruned through different pruning techniques. Error bars correspond to one standard deviation on the number of weights that remain unpruned across the six experimental seeds used in this work.
Of the pruning techniques tested, global unstructured pruning is the most consistent one in identifying similar sub-networks within a larger network across a variety of image classification tasks, if we exclude the random pruning technique, which are deterministically random. As a sanity check, random pruning techniques (both structured and unstructured) consistently prune the same (random) set of weights across all 5 tasks when the seed is kept constant, because, indeed, the choice of weights to prune is solely determined by the random seed and it is not task-dependent. A weight is then either always dropped (in bin 0) or never dropped (in bin 5).

\begin{figure}
\centering
  \subfloat[Number of weights per layer in the prêt-à-porter ticket]{\includegraphics[width=0.42\textwidth]{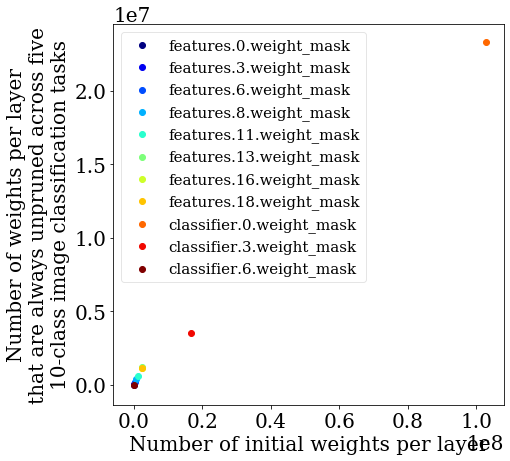}}
  \hspace{2em}
  \subfloat[Fraction of weights per layer in the prêt-à-porter ticket]{\includegraphics[width=0.42\textwidth]{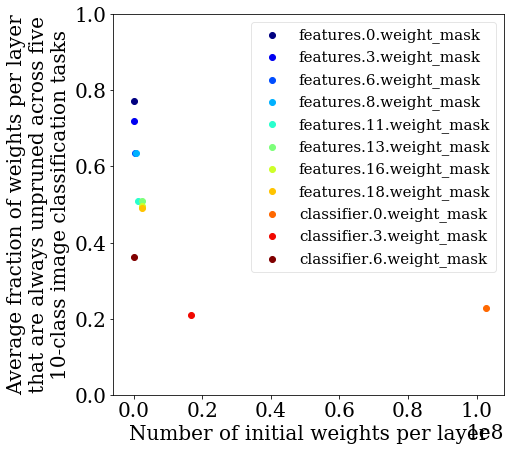}}
     \caption{Prêt-à-porter lottery ticket composition for VGG11.}\label{fig:intersection_vgg_nf}
\end{figure}

\begin{figure}
\centering
  \subfloat[Number of weights per layer in the prêt-à-porter ticket]{\includegraphics[width=0.47\textwidth]{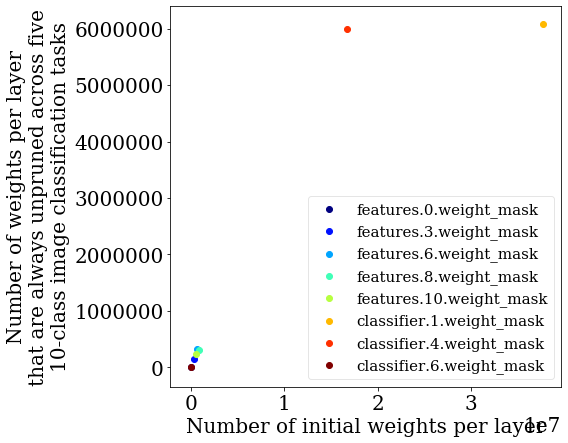}}
  \hspace{2em}
  \subfloat[Fraction of weights per layer in the prêt-à-porter ticket]{\includegraphics[width=0.42\textwidth]{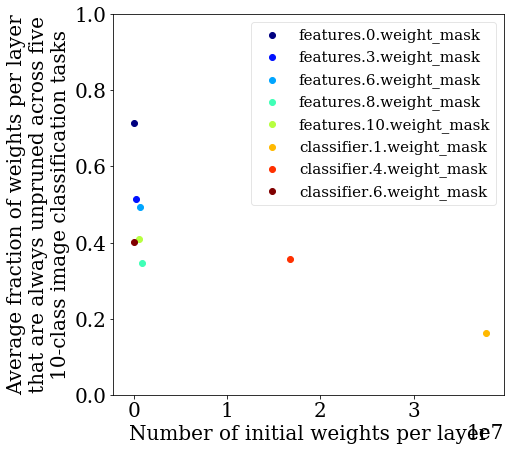}}
     \caption{Prêt-à-porter lottery ticket composition for AlexNet.}\label{fig:intersection_alex_nf}
\end{figure}

\begin{figure}
\centering
  \subfloat[Number of weights per layer in the prêt-à-porter ticket]{\includegraphics[width=0.47\textwidth]{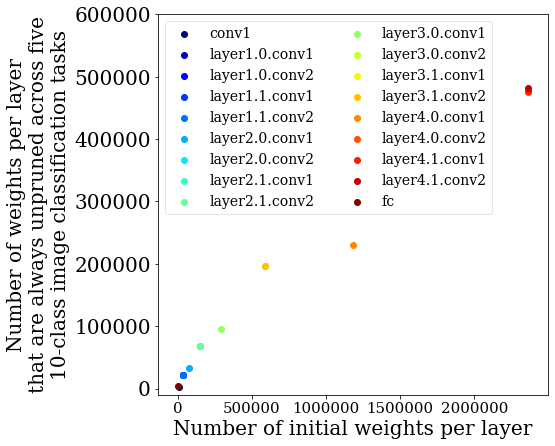}}
  \hspace{2em}
  \subfloat[Fraction of weights per layer in the prêt-à-porter ticket]{\includegraphics[width=0.42\textwidth]{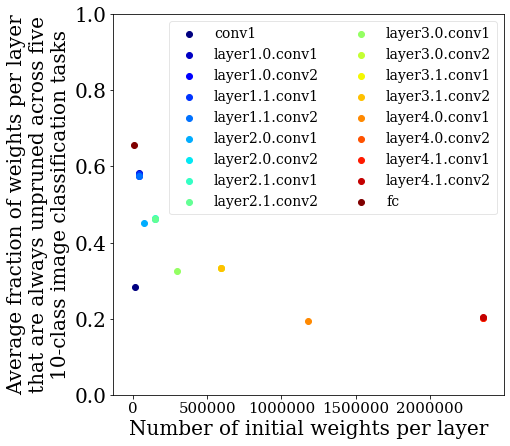}}
     \caption{Prêt-à-porter lottery ticket composition for ResNet18.}\label{fig:intersection_resnet_nf}
\end{figure}

\begin{figure}
\centering
  \subfloat[Number of weights per layer in the prêt-à-porter ticket]{\includegraphics[width=0.42\textwidth]{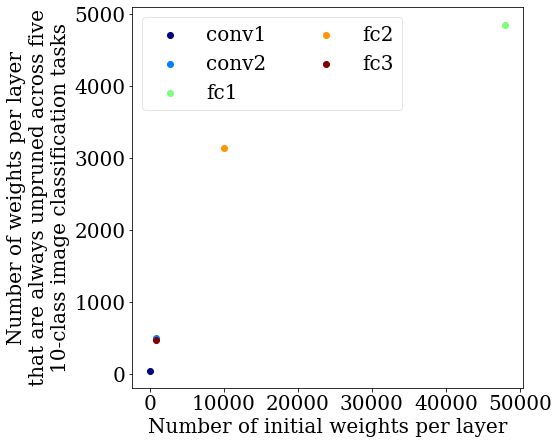}}
  \hspace{2em}
  \subfloat[Fraction of weights per layer in the prêt-à-porter ticket]{\includegraphics[width=0.42\textwidth]{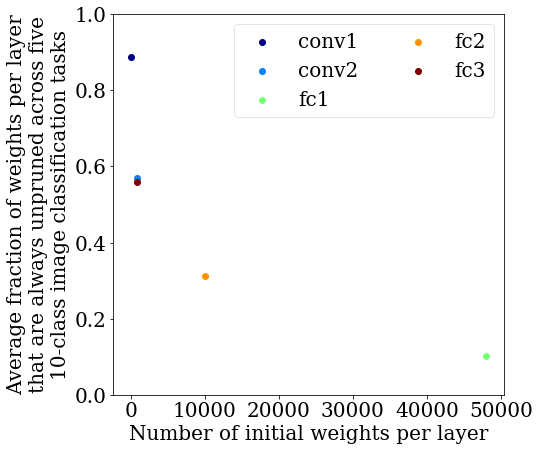}}
     \caption{Prêt-à-porter lottery ticket composition for LeNet.}\label{fig:intersection_lenet_nf}
\end{figure}

Fig.~\ref{fig:intersection_vgg_nf} shows the number and fraction of weights in each VGG11 layer that are selected by consensus to create the prêt-à-porter lottery ticket. The fraction of weights that is constantly unpruned across all five tasks rapidly declines as a function of the layer size; nonetheless, the largest initial layers remain largest in size even in the prêt-à-porter ticket. Similar plots are available for AlexNet in Fig.~\ref{fig:intersection_alex_nf}, for ResNet18 in Fig.~\ref{fig:intersection_resnet_nf}, and for Lenet in Fig.~\ref{fig:intersection_lenet_nf}.

\begin{figure}
    \centering
    \includegraphics[width=\textwidth]{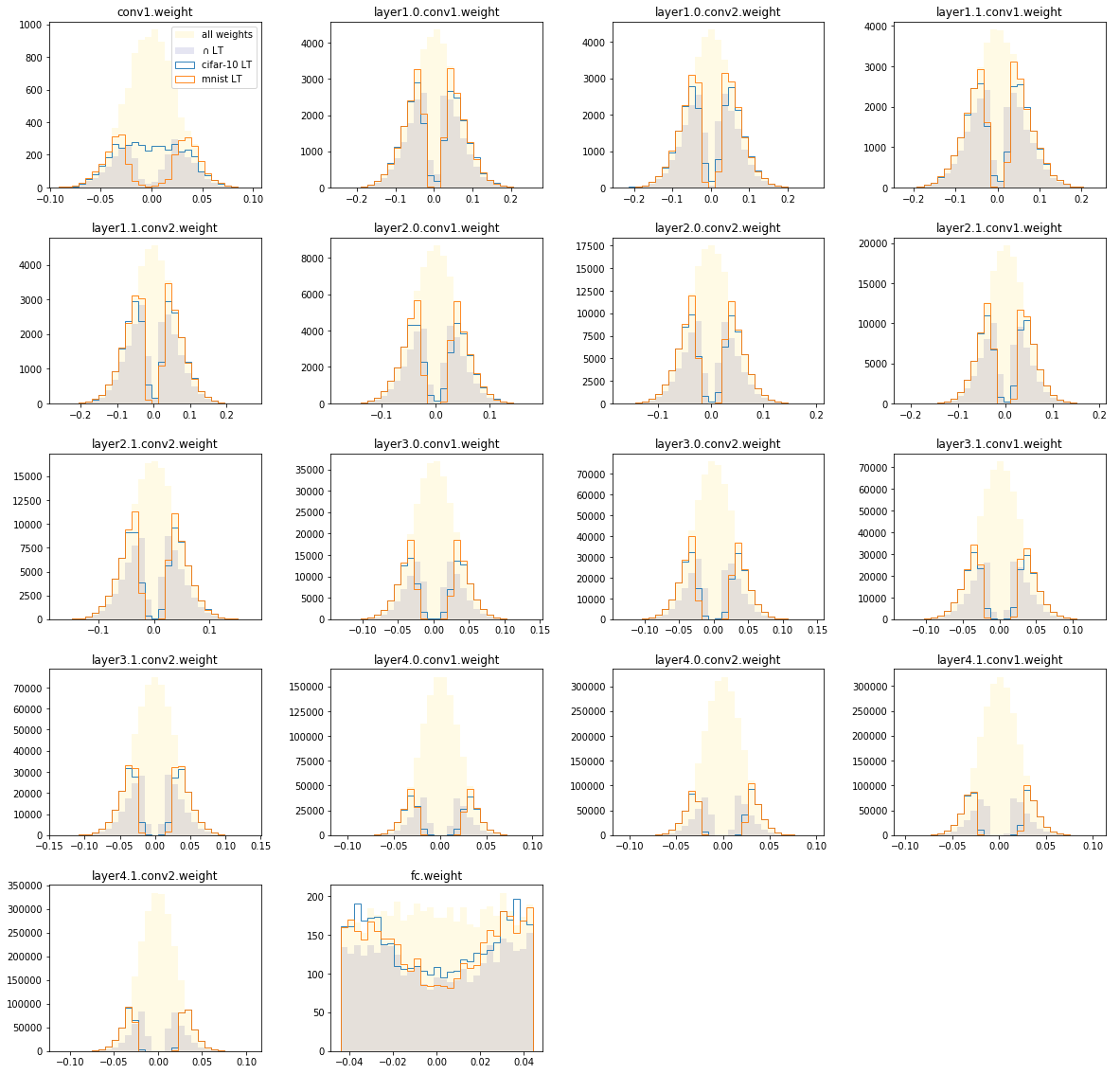}
     \caption{Initialization values of weights in a ResNet18 prêt-à-porter lottery ticket, compared to those that make up a bespoke LT of similar sparsity sourced on CIFAR-10, and those that make up a bespoke LT of similar sparsity sourced on MNIST. For comparison, the full distribution of initial weights is shown as a shaded gold histogram.}
    \label{fig:pap_init}
\end{figure}

\begin{figure}
    \centering
    \includegraphics[width=0.8\textwidth]{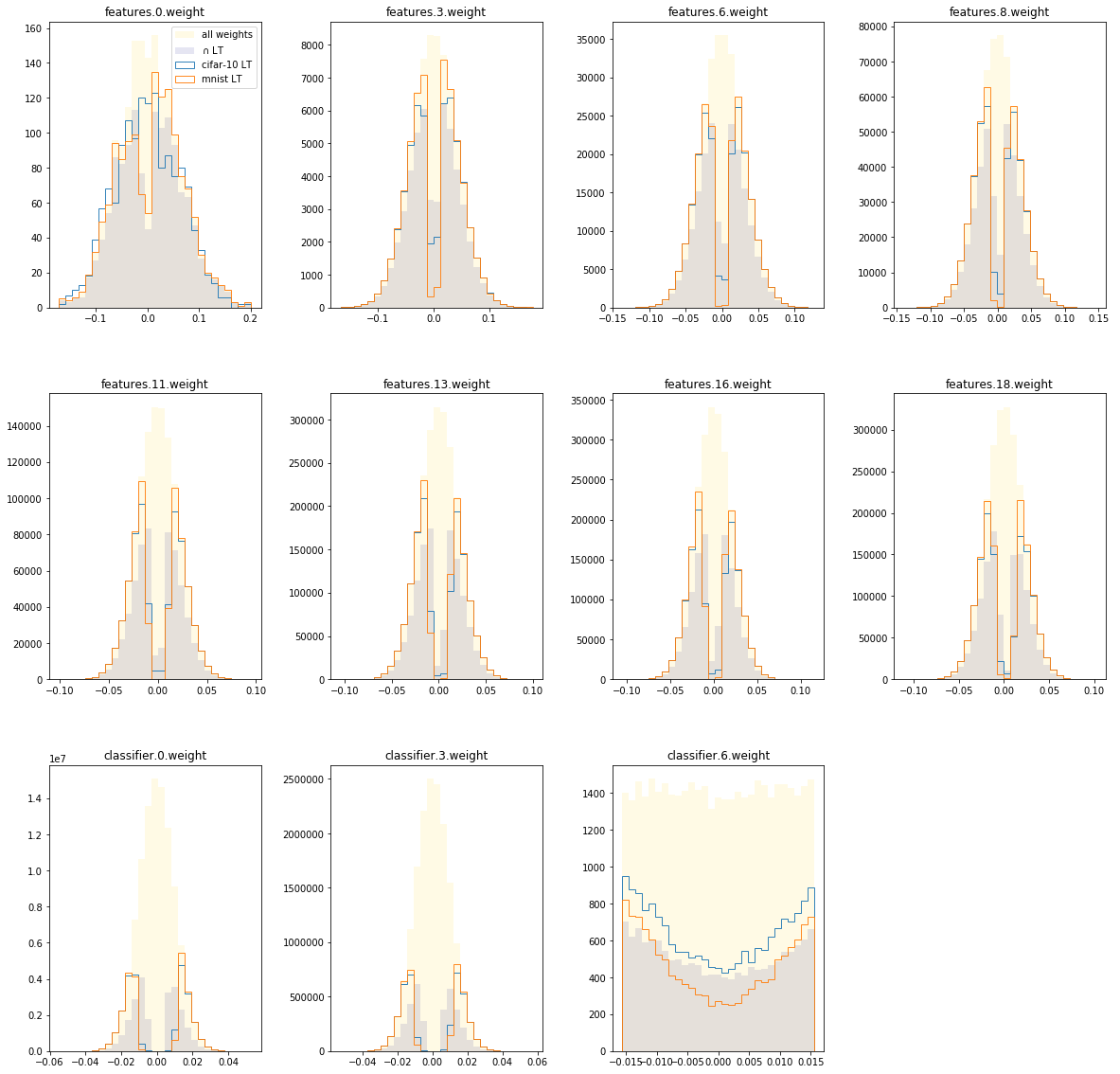}
     \caption{Initialization values of weights in a VGG11 prêt-à-porter lottery ticket, compared to those that make up a bespoke LT of similar sparsity sourced on CIFAR-10, and those that make up a bespoke LT of similar sparsity sourced on MNIST. For comparison, the full distribution of initial weights is shown as a shaded gold histogram.}
    \label{fig:pap_init_vgg}
\end{figure}

\begin{figure}
    \centering
    \includegraphics[width=\textwidth]{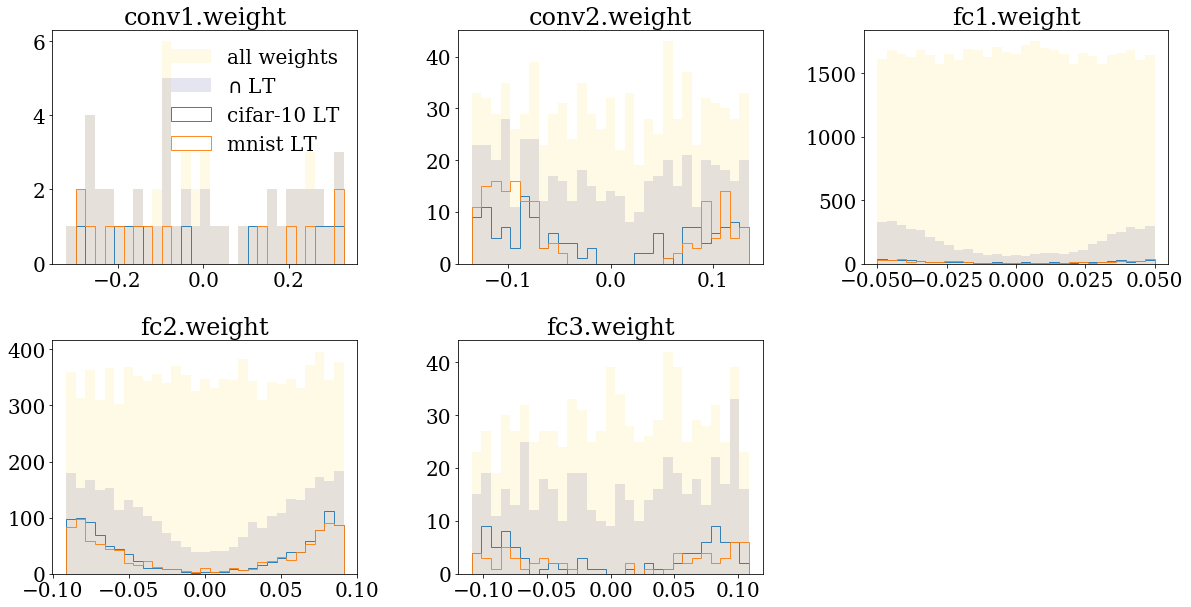}
     \caption{Initialization values of weights in a LeNet prêt-à-porter lottery ticket, compared to those that make up a bespoke LT of similar sparsity sourced on CIFAR-10, and those that make up a bespoke LT of similar sparsity sourced on MNIST. For comparison, the full distribution of initial weights is shown as a shaded gold histogram.}
    \label{fig:pap_init_lenet}
\end{figure}

Fig.~\ref{fig:pap_init}, ~\ref{fig:pap_init_vgg}, and~\ref{fig:pap_init_lenet} show the distribution of weights at initialization in ResNet18, VGG11, and LeNet networks, compared to the weights at initialization in the following subsets: parameters in the prêt-à-porter lottery ticket; parameters in a CIFAR-10 bespoke ticket; parameters in an MNIST bespoke ticket. All tickets were obtained via global unstructured pruning. 

\begin{figure}
\centering
  \subfloat[ResNet18]{\includegraphics[width=0.9\textwidth]{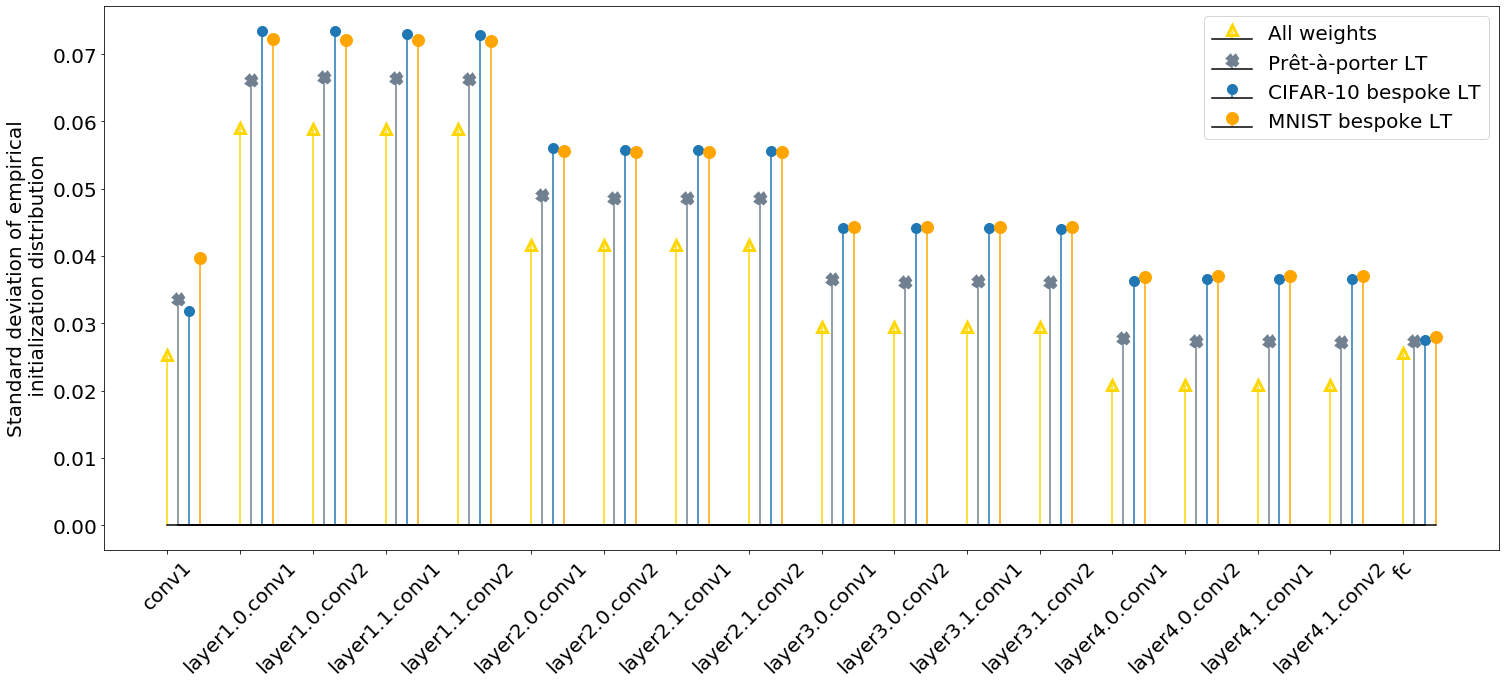}}\\
  \subfloat[AlexNet]{\includegraphics[width=0.75\textwidth]{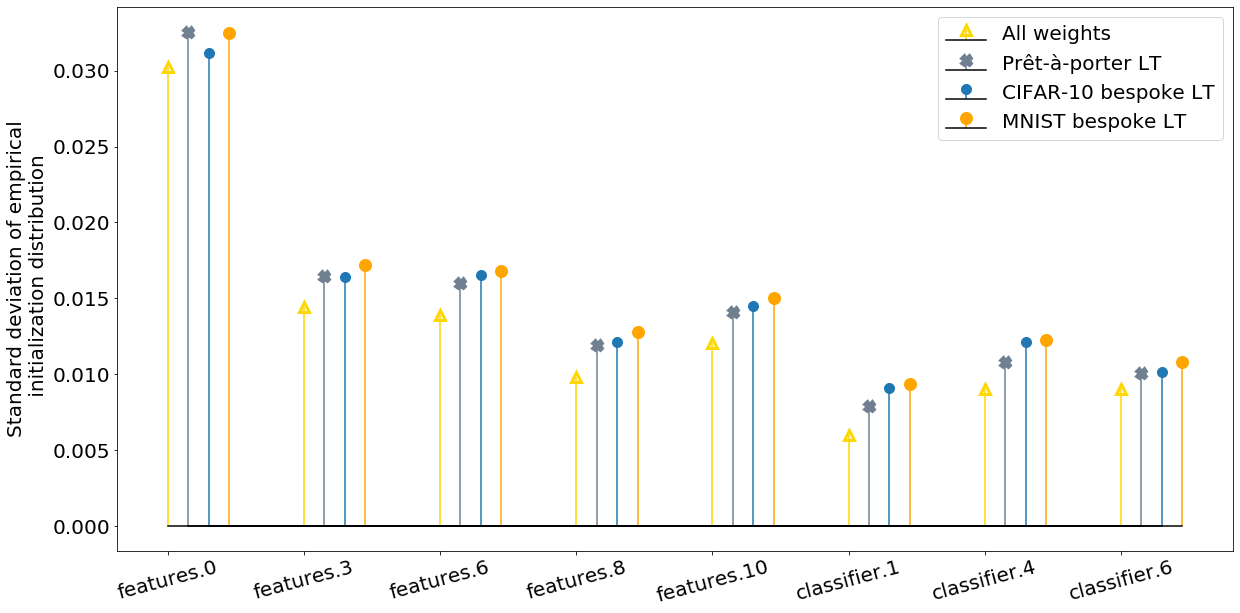}}\\
  \subfloat[LeNet]{\includegraphics[width=0.4\textwidth]{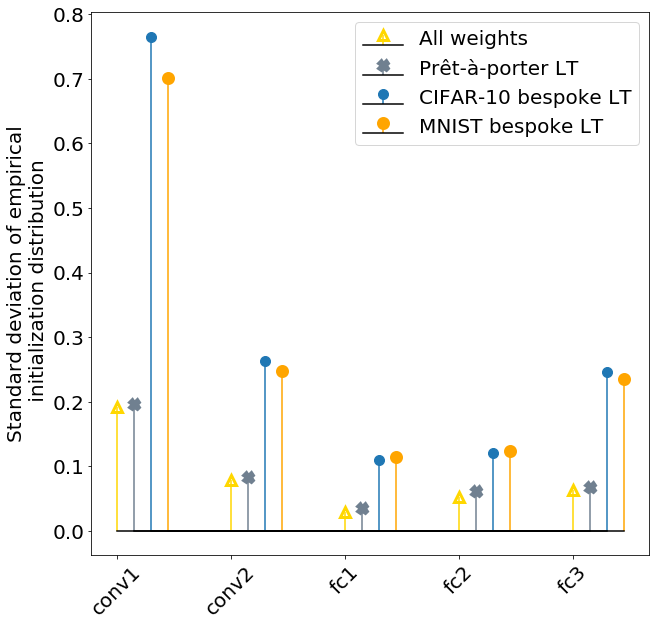}}
  \hspace{0.4cm}
  \subfloat[VGG11]{\includegraphics[width=0.5\textwidth]{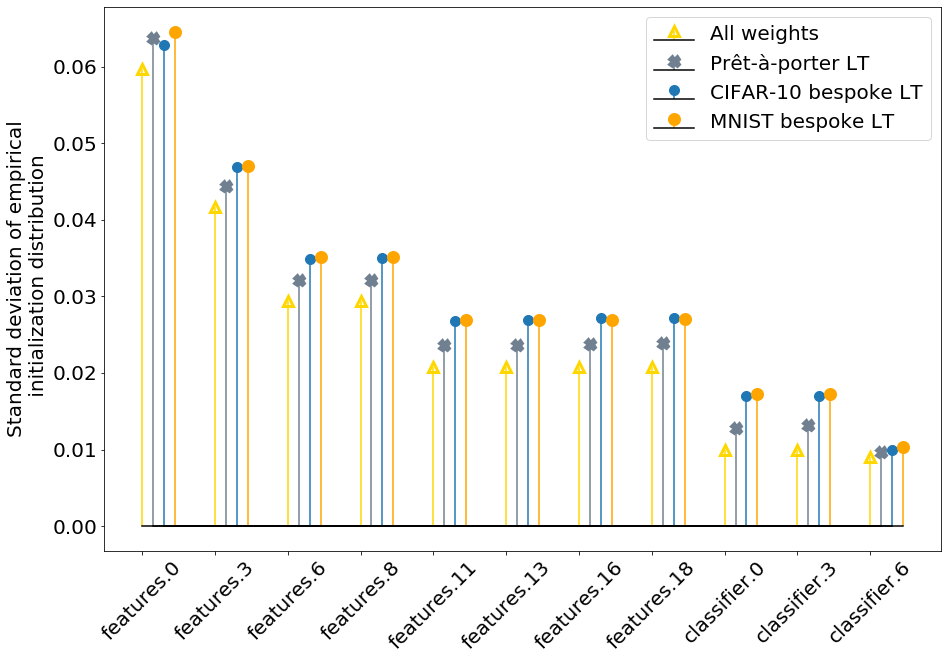}}\\
     \caption{Standard deviation of weight at initialization in: the whole layer, the fraction of weights in the layer that make up the prêt-à-porter ticket, the fraction of weights that makes up the bespoke CIFAR-10 ticket, and the fraction of weights that makes up the bespoke MNIST ticket.}\label{fig:init_stems}
\end{figure}

The distributions of initial values of the parameters that make up the prêt-à-porter lottery tickets show a stronger tendency towards smaller magnitude weights, compared to bespoke tickets. This is more explicitly quantified in Fig.~\ref{fig:init_stems}, where, for each architecture considered in this work, we compute the average empirical standard deviation of weights at initialization in each layer. In gold, we mark the empirical standard deviation of the full weight distribution at initialization prior to any pruning; in grey, we mark the standard deviation of our prêt-à-porter ticket at initialization; in blue, for the CIFAR-10 bespoke ticket; in orange, for the MNIST bespoke ticket.
This larger distribution variance at initialization observed in bespoke tickets is due to the explicit choice of removing low magnitude weights at each iteration of pruning, over 20 pruning iterations, in the LT-finding algorithm that selects bespoke tickets. In the case of prêt-à-porter LTs, instead, the algorithm we propose only explicitly removes the lowest magnitude weights over the two magnitude-based pruning iterations applied to the $N=5$ source tickets. The remainder of the sparsification procedure does not explicitly select out parameters that correlate with low absolute value at initialization.

In both bespoke and prêt-à-porter LTs, weights that make up lottery tickets are distributed according to a symmetric bimodal distribution at initialization.

\section{Lottery Ticket performance at initialization}
\label{app:atinit}
\begin{sidewaysfigure}
    \centering
    \includegraphics[width=\textwidth]{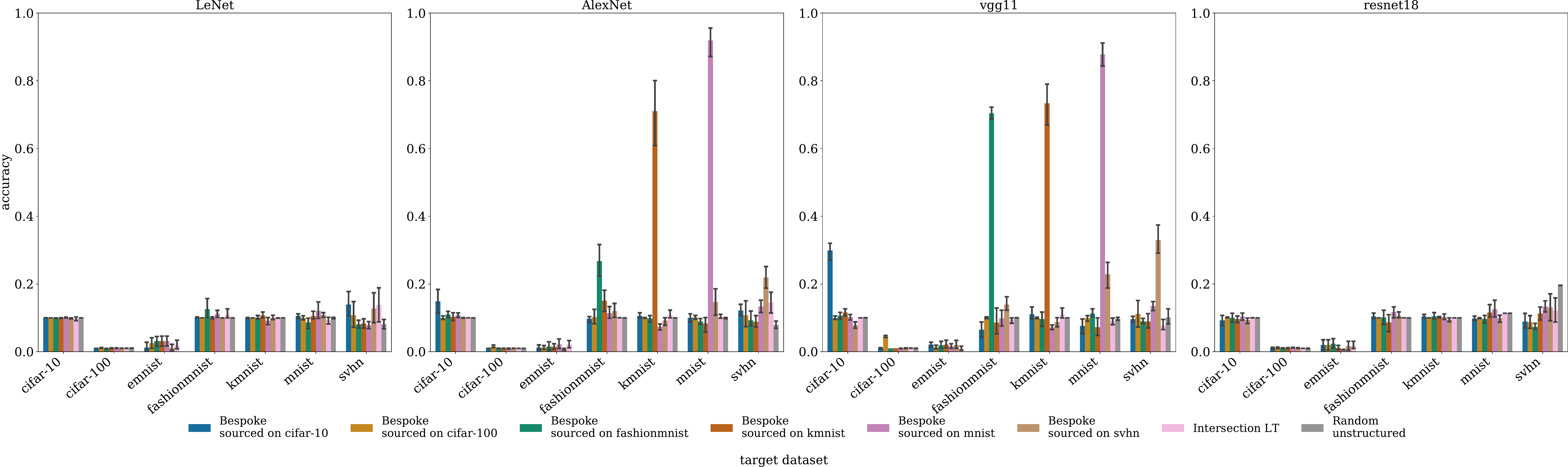}
    \caption{Performance of bespoke and prêt-à-porter lottery tickets obtained through global unstructured pruning from different base architectures (see each panel) at initialization. The source of the tickets can be found in the legend. The x-axis labels the target dataset on which the tickets were evaluated. These tickets correspond to the ones that were then retrained and evaluated to generate the results in Fig.~\ref{fig:across_tasks}.}
    \label{fig:atinit}
\end{sidewaysfigure}

In the spirit of prior work on Lottery Tickets and \textit{supermasks}~\cite{Zhou2019-xh}, we evaluate the performance of bespoke and prêt-à-porter lottery tickets at initialization, prior to any training taking place. We are unsuccessful at extracting tickets with better-than-random performance for any of the task-architecture combinations considered in this work, with the exception of bespoke tickets sourced on large, over-parametrized, and often parameter-inefficient networks such as AlexNet and VGG11 (Fig.~\ref{fig:atinit}). This allows us to notice that, while bespoke tickets are flexible enough to be successfully retrained on target tasks, they do have source dataset-specific properties that allow them to align with a good solution on their source domain even at initialization. This specialization, however, doesn't prevent them from transferring, and does not result in large performance gains post-training.

Due to problems encountered with job submission, certain lottery ticket / target dataset pairs are not present for all six seeds. For VGG11, one seed is present for ``Bespoke sourced on fashionmnist" and ``Bespoke sourced on fashionmnist" evaluated on CIFAR-100. For ResNet18, Only five of six seeds are available for ResNet18 ``Bespoke sourced on svhn" for MNIST and EMNIST and for ``Intersection LT" for CIFAR-100; four of six seeds are available for ``Bespoke sourced on svhn" all other target datasets.  Results for random unstructured on EMNIST are not available on ResNet18.

\section{Lottery ticket performance on held out datasets}

\begin{figure}
    \centering
    \includegraphics[width=\textwidth]{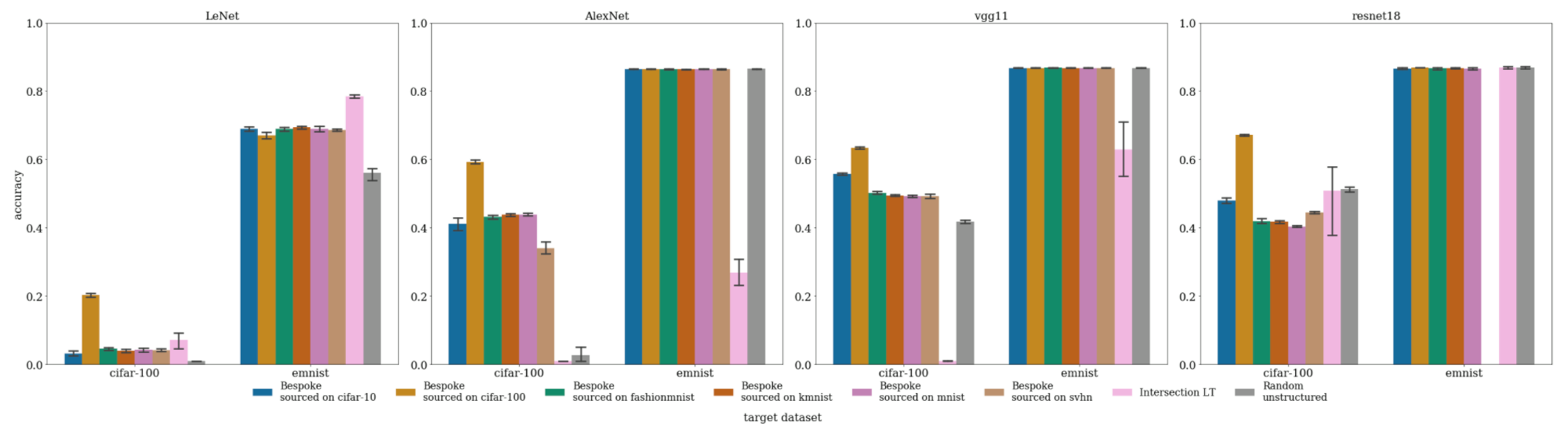}
    \caption{Performance of bespoke and prêt-à-porter lottery tickets obtained through global unstructured pruning from different base architectures (see each panel). The source of the tickets can be found in the legend. The x-axis labels the target dataset on which the tickets were evaluated. These tickets correspond to the ones that were then retrained and evaluated to generate the results in Fig.~\ref{fig:across_tasks}.}
    \label{fig:external_eval}
 \end{figure}

We report the performance for the bespoke lottery tickets and prêt-à-porter lottery tickets on two additional tasks not used in training, CIFAR-100 and EMNIST, Fig.~\ref{fig:external_eval}.  Due to problems encountered with job submission, results are reported for two out of six seeds for ``Bespoke sourced on cifar-100" on EMNIST and are missing for ``Bespoke sourced on svhn" on EMNIST.

\section{Further experimental details for reproducibility}
\subsection{Models}
The models trained and pruned in this work correspond to the following base architectures:
\begin{itemize}
    \item LeNet, with two convolutional layers and three fully connected layers. The convolutional layers perform $3\times3$ convolutions with 6 and 16 output channels, respectively. The fully connected layers transform the hidden representation from 400-dimensional to 120-d, 84-d, and 10-d at the output layer, for 10 class classification problems. Wherever the dataset requires a larger number of outputs, the final layer is replaced with a fully-connected layer of dimensions matching the number of output classes in the task. Rectified linear units are used as activations throughout. Two MaxPool operations are executed after the first two non-linear activations.
    \item AlexNet, as implemented in \texttt{torchvision}
    \item VGG11, as implemented in \texttt{torchvision}
    \item ResNet18, as implemented in \texttt{torchvision}
\end{itemize}
All networks are trained from scratch, with each layer initialized with the default PyTorch initialization schema.

\subsection{Pruning}
Given a tensor $\tT$, representing the unpruned parameters in a layer:
\begin{itemize}
    \item \texttt{l1\_unstructured} pruning removes the entries $\left\{\etT_k \ : k_\mathrm{th}\left(\{\etT_i\}_{i=1}^{|\tT|} \ \middle| \ ||\cdot||_1\right)<0.2\right\}$, where $k_\mathrm{th}$ is the rank of the $k^\mathrm{th}$ element of $\{\etT_i\}_{i=1}^{|\tT|}$, according to the $L_1$-based ranking, and $t=0.2$ is the pruning fraction per iteration used in this work;
    \item \texttt{l1\_structured} pruning removes all entries in channel $k$ along dimension $d$ defined as $\left\{ \tT_{[...,\ k, ...]} \ : \ k_\mathrm{th}\left( \{\tT_{[...,\ i, ...]}\}_{i=1}^{|d|} \middle| \ ||\cdot||_1 \right) < 0.2 \right\}$, where $k_\mathrm{th}$ is the rank of the $k^\mathrm{th}$ channel of $\tT$ along direction $d$ according to the $L_1$-based ranking, and $t=0.2$ is the pruning fraction per iteration used in this work;
    \item \texttt{l2\_structured} pruning removes all entries in channel $k$ along dimension $d$ defined as $\left\{ \tT_{[...,\ k, ...]} \ : \ k_\mathrm{th}\left( \{\tT_{[...,\ i, ...]}\}_{i=1}^{|d|} \middle| \ ||\cdot||_2 \right) < 0.2 \right\}$, where $k_\mathrm{th}$ is the rank of the $k^\mathrm{th}$ channel of $\tT$ along direction $d$ according to the $L_2$-based ranking, and $t=0.2$ is the pruning fraction per iteration used in this work;
    \item \texttt{linfty\_structured} pruning removes all entries in channel $k$ along dimension $d$ defined as $\left\{ \tT_{[...,\ k, ...]} \ : \ k_\mathrm{th}\left( \{\tT_{[...,\ i, ...]}\}_{i=1}^{|d|} \middle| \ ||\cdot||_{\infty} \right) < 0.2 \right\}$, where $k_\mathrm{th}$ is the rank of the $k^\mathrm{th}$ channel of $\tT$ along direction $d$ according to the $L_\infty$-based ranking, and $t=0.2$ is the pruning fraction per iteration used in this work;
    \item \texttt{random\_structured} pruning removes all entries in channel $k$ along dimension $d$ chosen at random: $\left\{ \tT_{[...,\ k, ...]} \ : \ k = \texttt{randint}(1, |d|, 0.2 * |d|)\right\}$, where $t=0.2$ is the pruning fraction per iteration used in this work;
    \item \texttt{random\_unstructured} pruning removes entries $\left\{\etT_k \ : k = \texttt{randint}(1, |\tT|, 0.2 * |\tT|) \right\}$, where $t=0.2$ is the pruning fraction per iteration used in this work.
\end{itemize}

Given a tensor $\tU$, representing the unpruned parameters in a set of layers:
\begin{itemize}
    \item \texttt{global\_unstructured} pruning removes the entries $\left\{\etU_k \ : k_\mathrm{th}\left(\{\etU_i\}_{i=1}^{|\tU|} \ \middle| \ ||\cdot||_1\right)<0.2\right\}$, where $k_\mathrm{th}$ is the rank of the $k^\mathrm{th}$ element of $\{\etU_i\}_{i=1}^{|\tU|}$, according to the $L_1$-based ranking, and $t=0.2$ is the pruning fraction per iteration used in this work;
\end{itemize}

\subsection{Prêt-à-porter lottery ticket finding algorithm}

Prêt-à-porter lottery ticket finding is elaborated upon in Alg.~\ref{pptalg}.

\begin{algorithm}[]
\SetAlgoLined
\KwData{Set of datasets $\mathcal{S}$ on which to source ticket, a network with weights $\theta$, pruning iterations $T$, epochs for iteration $E$.}
\KwResult{A consensus mask $\mathbb{M}$}
\BlankLine
 $\mathbb{M} = I_{\vert \theta \vert}$ \;
 \For{dataset $S \in \mathcal{S}$}{

  $\theta_{S} = \theta$ \;

  \For{$t\leftarrow 1$ \KwTo $T$}{
     \For{$e\leftarrow 1$ \KwTo $E$}{

       $\theta_{\mathsf{trained}} = \mathsf{TrainEpoch}(\theta_{S})$ \;
       
      }

   $\theta_{\mathsf{pruned}} = \mathsf{Prune}(\theta_{\mathsf{trained}})$ \;
   $\theta_{S} = \mathsf{Reinit}(\theta_{\mathsf{pruned}})$ \;
   
  }
  
  $\mathbb{M} = \mathbb{M} \cap \mathsf{GetMask}(\theta_{S})$ \;
  
 }
 
 return $\mathbb{M}$ \;

 \caption{Prêt-à-porter (consensus) lottery ticket finding algorithm}
 \label{pptalg}
\end{algorithm}

\subsection{Data}

Seven datasets are used in experiments: MNIST, KMNIST, FashionMNIST, CIFAR-10, CIFAR-100, SVHN. Table~\ref{table:dataset-details} shows the sizes of each dataset, prior to training set splitting. The training set is, in fact, randomly split into a proper training set (80\% of the original training set) and a validation set (20\% of the original training set), using \texttt{torch.utils.data.random\_split}. The random split is controlled by the same experimental seed that makes the rest of the experiments reproducible. 

All images are normalized to the training set statistics. At train time, images are subjected to data-augmentation via a horizontal flip and a random crop. All images are resized to fit the expect input size of the model they are being processed with. In practice, when images are fed into a ResNet, VGG, or AlexNet model, they are resizes to size $(224, 224, 3)$. When they are processed by a LeNet architecture, they are reshaped to size $(28, 28, 1)$.
 
\begin{table}[]
\centering
\caption{Specifics of Datasets used in Experiments}
\begin{tabular}{@{}lllll@{}}
\toprule
Dataset         & Training Examples & Testing Examples            & Citation                                       \\ \midrule
MNIST           & 60,000            & 10,000           & \cite{mnist:1994}             \\
Kuzushiji-MNIST & 60,000            & 10,000           & \cite{clanuwat2018deep, kmnist}       \\
Fashion-MNIST   & 60,000            & 10,000           & \cite{fashionmnist}           \\
CIFAR-10        & 60,000            & 10,000           & \cite{Krizhevsky2009-fx} \\
CIFAR-100       & 60,000            & 10,000           & \cite{Krizhevsky2009-fx} \\
SVHN            & 73,257            & 26,032           & \cite{netzer2011reading}      \\ \bottomrule
\end{tabular}
\label{table:dataset-details}
\end{table}

\end{document}